\documentclass[twoside,leqno,twocolumn]{article}

\usepackage[letterpaper]{geometry}

\usepackage{amsthm, amsmath, amssymb}
\usepackage{subfig}
\usepackage{makecell}
\usepackage{ltexpprt}
\usepackage{hyperref}
\usepackage{graphicx}
\usepackage{algorithm}
\usepackage{algpseudocode}
\newcommand{\email}[1]{\href{mailto:#1}{\nolinkurl{#1}}}

\usepackage{hyperref} % for clickable email addresses

\begin{document}

\newcommand\relatedversion{}

\title{\Large HeteroMILE: a Multi-Level Graph Representation Learning Framework for Heterogeneous Graphs\relatedversion}
\author{Yue Zhang \thanks{The Ohio State University
(\email{zhang.8016@osu.edu}).}\and Yuntian He \thanks{The Ohio State University
(\email{he.1773@osu.edu}).}
\and Saket Gurukar \thanks{The Ohio State University
(\email{gurukar.1@osu.edu}).}
\and Srinivasan Parthasarathy\thanks{The Ohio State University
(\email{srini@cse.ohio-state.edu}).}}
\date{}

\maketitle

% Copyright Statement
% When submitting your final paper to a SIAM proceedings, it is requested that you include
% the appropriate copyright in the footer of the paper.  The copyright added should be
% consistent with the copyright selected on the copyright form submitted with the paper.
% Please note that "20XX" should be changed to the year of the meeting.

% Default Copyright Statement
\fancyfoot[R]{\scriptsize{Copyright \textcopyright\ 2024 by SIAM\\
Unauthorized reproduction of this article is prohibited}}

% Depending on which copyright you agree to when you sign the copyright form, the copyright
% can be changed to one of the following after commenting out the default copyright statement
% above.

%\fancyfoot[R]{\scriptsize{Copyright \textcopyright\ 20XX\\
%Copyright for this paper is retained by authors}}

%\fancyfoot[R]{\scriptsize{Copyright \textcopyright\ 20XX\\
%Copyright retained by principal author's organization}}

%\pagenumbering{arabic}
%\setcounter{page}{1}%Leave this line commented out.

\begin{abstract} \small\baselineskip=9pt   Heterogeneous graphs are ubiquitous in real-world applications because they can represent various relationships between different types of entities. Therefore, learning embeddings in such graphs is a critical problem in graph machine learning. However, existing solutions for this problem fail to scale to large heterogeneous graphs due to their high computational complexity. To address this issue, we propose a Multi-Level Embedding framework of nodes on a heterogeneous graph (HeteroMILE) - a generic methodology that allows contemporary graph embedding methods to scale to large graphs. HeteroMILE repeatedly coarsens the large-sized graph into a smaller size while preserving the backbone structure of the graph before embedding it, effectively reducing the computational cost by avoiding time-consuming processing operations. It then refines the coarsened embedding to the original graph using a heterogeneous graph convolution neural network. We evaluate our approach using several popular heterogeneous graph datasets. The experimental results show that HeteroMILE can substantially reduce computational time (approximately 20x speedup) and generate an embedding of better quality for link prediction and node classification.
\end{abstract}

\section{Introduction}
Graphs serve as a versatile representation for capturing relationships across diverse domains, including social networks, biological systems, and information networks. In these applications, entities and their relationships are often in different types, leading to the concept of heterogeneous graphs. Heterogeneous graphs, characterized by nodes and edges with diverse types and attributes, offer a more comprehensive representation of complex systems \cite{sun2011pathsim}. Recently, graph embedding has gained significant attention as a means to capture both the structural and content-related aspects of graphs in a lower-dimensional vector space. This representation effectively preserves essential graph properties and information.  Graph embedding finds applicability across various types of graphs, including homogeneous and heterogeneous graphs, and has been leveraged in diverse domains like social network analysis, bioinformatics, and natural language processing \cite{maruhashi2011multiaspectforensics}. However, existing graph embedding techniques face challenges when it comes to scalability. Particularly, the embedding techniques that rely on random walks, such as metapath2vec \cite{dong2017metapath2vec} demand considerable CPU time to generate a sufficient number of walks for training the embedding model. This computationally intensive process hinders their scalability, restricting their applicability to large-scale graphs.

In recent years, improvements in the scalability of homogeneous graph embedding have been extensively studied. A popular methodology adopted by several prior studies \cite{liang2021mile, chen2018harp, fu2019learning} is the multi-level framework.
% There are several approaches to improve the scalability of graph embedding techniques. For example, \citet{liang2021mile} proposed a multi-level embedding framework MILE for homogeneous graphs. 
The core concept of their approach involves iteratively reducing the complexity of the original graph through coarsening, followed by the application of a pre-existing embedding method to the simplified graph. The resulting embeddings are further refined using graph neural networks (GNN). Through extensive experimentation, they have shown that this approach enhances scalability without sacrificing the performance of the embeddings.

% Other methods such as HARP \cite{chen2018harp} have similar strategy as MILE with only minor changes in coarsening part.
% The advantages of MILE is that ....
However, these approaches are not trivial to extend to heterogeneous graphs, as one must carefully account for node heterogeneity in both the coarsening and refinement procedures.  
To our understanding, this is the initial endeavor to tackle this issue. Specifically,  we propose a Multi-Level Embedding framework on a Heterogeneous graph (HeteroMILE). HeteroMILE consists of three major steps: Firstly, we employ various matching strategies to iteratively reduce the size of the large graph, resulting in a smaller coarsened graph. Secondly, we apply well-established heterogeneous graph embedding techniques \cite{dong2017metapath2vec,cen2019representation}to compute embeddings on this coarsened graph. Lastly, we introduce a refinement model that utilizes a heterogeneous graph convolution network to enhance the embeddings, refining them from the smaller-sized graph back to the original-sized graph. We demonstrated the viability of the HeteroMILE framework on several popular datasets: AcademicII, DBLP, IMDB, and OGB\_MAG. Our current results show that HeteroMILE can significantly reduce the computational time and improve link prediction and node classification accuracy. 
The primary contributions of this research can be outlined as follows:

\begin{itemize}
\item In order to address the scalability issue in heterogeneous graph embedding, we propose HeteroMILE which is a generalizable multi-level framework improving the efficiency of well-established heterogeneous graph embedding methods. Unlike existing approaches, HeteroMILE is not simply reliant on advanced computing resources. We demonstrate the effectiveness and efficiency of HeteroMILE using Metapath2Vec and GATNE\cite{dong2017metapath2vec,cen2019representation}, popular heterogeneous graph embedding strategies, as a proof-of-concept base embedding strategy. 
 % \item We propose HeteroMILE, a multi-level embedding framework on heterogeneous graphs. We implemented several coarsen strategies to reduce the size of the graph while preserving the structure information and then utilized the embedding techniques on the small-size graph and thirdly refine the coarsened embeddings to the original embeddings using a refinement model.

 \item We propose two novel coarsening algorithms that adapt to the heterogeneous context - an exact Jaccard Similarity based approach and a more efficient approximate approach based on Locality-Sensitive Hashing (LSH) for node matching. We explored different experimental designs for the two algorithms to improve the performance and speed up the coarsening process.

\item Our proposed approach involves utilizing a Heterogeneous Graph Convolutional Network \cite{yang2021interpretable} as a refinement model to enhance the embeddings obtained from the smaller-size graph, refining them to align with the initial graph.

\item We perform our experiments on diverse and real-world datasets of heterogeneous graphs. We used four heterogeneous graph datasets, including academic graphs and Internet movie databases. On the largest dataset, OGB\_MAG, which reflects the relationship between authors and institutions, papers, and fields, HeteroMILE is more than 20x faster than the baseline and offers improved performance to boot (up to a certain coarsening level).

\item We evaluate different parameters on HeteroMILE. The experimental results show a tradeoff between performance and computation cost, but also prove that HeteroMILE can work well in different parameter settings.

\end{itemize}

\section{Background and Motivation}

\subsection{Heterogeneous Graph Embedding} % Metapath2vec
%hetero graph emb 
% bring to 2.2

Several algorithms have been proposed for learning node representations in homogeneous networks ~\cite{grover2016node2vec, perozzi2014deepwalk}. However, applying these methods to heterogeneous graphs directly is not feasible due to the inherent heterogeneity present in the graph data. Specifically, heterogeneous graphs have different relationships between nodes and may require specific efforts for information fusion across different node and edge spaces.

Recent studies have focused on heterogeneous graph embeddings. Several methods learn node representations by leveraging heterogeneous substructures and semantics. A representative work using this idea is metapath2vec \cite{dong2017metapath2vec}, which generates the embeddings by performing random walks using predefined \textit{meta-paths} and training a skip-gram model \cite{guthrie2006closer} with these sequences. Meta-path is a defined sampling scheme to capture a specific semantic between different node and edge types (e.g., "author-paper-author" denotes a co-authorship in a citation network). Another representative embedding method is GATNE \cite{cen2019representation} which also performs random works based on \textit{meta-paths} but considers both graph structure and temporal information. It employs graph attention networks and attention mechanisms to capture the structure dependencies and temporal dynamics of the network. Other studies learn heterogeneous embeddings by preserving other substructures, such as links \cite{chen2018pme, tang2015pte, zhang2018scalable} and subgraphs \cite{zhang2018metagraph2vec, tu2018structural}. In addition, another group of studies leveraged the rich attributes and built heterogeneous graph neural networks (HGNNs) \cite{wang2019heterogeneous, zhang2019heterogeneous} for performance improvement.

Despite the promising performance of these methods, their applicability is limited to large-scale heterogeneous graphs, owing to their complexity. For example, they require exceptional time for sampling (random-walk-based methods) or training (HGNN-based methods). This motivates our study of HeteroMILE, which aims to help contemporary heterogeneous graph embedding techniques scale up to large networks.

% % Metapath2vec \cite{dong2017metapath2vec} is a representation learning model which generate the embedding of the graph nodes by a set of meta-paths \cite{liu2011metapath}. A meta-path is a sequence of node types, where consists a sequence of relation defined between different node types in the graph. For example, a meta-path in a co-authorship network might be "author-paper-author" to represent the relationship between two authors that have co-authored a paper.

% % The main idea behind Metapath2vec is to use these meta-paths to define a context for the nodes, and then use this context to learn a representation of the nodes. This is done by training a skip-gram \cite{guthrie2006closer} model on the nodes, where the context is defined by the meta-paths. The effectiveness of Metapath2vec has been evaluated on several real-world datasets, including a co-authorship network, a bibliographic network, and a protein-protein interaction network \cite{yu2014predicting, koh2012analyzing}. The resulting representations can be used for various tasks such as link prediction, node classification, and community detection \cite{lu2011link, bhagat2011node, fortunato2010community}. 

% % One of the key advantages of Metapath2vec is that it is able to capture higher-order relationships between nodes in the graph, which are not captured by traditional methods that only consider the immediate neighbors of a node. Additionally, Metapath2vec can handle graphs with multiple types of nodes and edges, and it can be easily extended to incorporate additional meta-paths.

\subsection{Scalable Graph Embedding} % Multi-Level Embedding
% scalable ml embedding
Recently, several approaches have been put forth to enhance the scalability of graph embedding techniques. Several studies have adopted a multi-level framework to improve scalability \cite{liang2021mile, he2022webmile, chen2018harp, akyildiz2020gosh}. % The essential idea of the multi-level framework is to first coarsen the input graph into a smaller one, then learn the embeddings on the coarsened graph and refine them. These methods learn representations efficiently, without compromising the quality of the results.

A multi-level embedding framework is an effective way to address the challenges of large graph embedding. The process begins with graph coarsening to decrease the size of the input graph, followed by learning the embeddings of the fine-grained graph based on the coarsened versions. By decreasing the graph size, this framework not only can improve the efficiency of the embedding process but also maintain high-order structural features for enhanced quality.

One recent work is MILE \cite{liang2021mile}, which is a framework for homogeneous graphs.
% It has three phases: coarsening the graph, generating the base embedding, and refining the embedding. 
% The process starts with the original graph $G$ (or $G_{0}$) and repeatedly coarsens m times it until the coarsest graph $G_{m}$ is obtained.
It first merges groups of nodes into supernodes and then combines the edges. MILE uses a combination of two methods to match nodes: Structural Equivalence Matching (SEM) and Normalized Heavy Edge Matching (NHEM). SEM matches nodes with the same neighbors, whereas NHEM matches unmatched nodes $u$ with their unmatched neighbor $v$ with the highest normalized weight edge ($u$, $v$). In addition, NHEM prioritizes nodes with more neighbors to collapse first to merge more nodes. Then, MILE generates the embeddings on the shrunk graph and projects the embeddings to the original graph.

% After merging the nodes, MILE applies a graph embedding method to the coarsest graph $G_{m}$ to learn the embeddings $E_{m}$. This is more efficient than directly embedding the original graph $G$ because the coarsest graph is smaller. This phase is treated as a black box, and can be applied using any embedding method.

%The last step of MILE is to pass the embedding to the refinement model. Given the embeddings on the coarsest graph ($E_{m}$) and the series of graphs {$G_{m}$, $G_{m-1}$,..., $G_{1}$, $G_{0}$}, MILE iteratively computes the embeddings of each graph and finally obtains $E_{0}$. This is achieved by training a GNN model such as GraphSAGE \cite{hamilton2017inductive} or GCN \cite{kipf2016semi}, which refines the embeddings $E_{i}$ to $E_{i-1}$. At each level, MILE projects the embeddings of supernodes in $G_{i}$ to their corresponding nodes in $G_{i-1}$ and then applies the trained model to refine the projected embeddings.

Another representative work is Graphzoom\cite{deng2019graphzoom}, which partitions the graph into multiple subgraphs using a spectral clustering algorithm and combines node attributes and the structure information to construct a fused graph. It then performs the embedding and maps it to the original graph. 

In addition to the multi-level framework, some studies \cite{zhu2019graphvite, lerer2019pytorch, zhang2020poster} used high-performance computing techniques, such as parallel computing, distributed computing, and GPU training, to resolve the scalability issue. Unfortunately, all of these studies were strictly designed for homogeneous graphs due to differences in the graph structure and the properties of nodes and edges. 
In homogeneous graphs, nodes, and edges typically have similar characteristics, making it easier to combine and merge the nodes. In contrast, heterogeneous graphs often contain nodes and edges with diverse properties, such as different types, attributes, or semantics. 
General matching methods of different node types would lead to a loss of contextual information, which causes less accurate results. Additionally, heterogeneous graphs often exhibit complex connectivity patterns between different types of nodes. The general merging approach might lose the patterns and ignore the intricate relationships and connections in the original graph. While existing studies fail to leverage heterogeneous data, our proposed HeteroMILE considers the specific node and edge types which is able to learn node embeddings on such networks while improving scalability.

% In addition to MILE, the researchers have also proposed some other embedding methods using the multi-level framework recently. One such method is HARP\cite{chen2018harp}. HARP has a similar approach to MILE which coarsens the input graph prior to embedding, while HARP learns $E_{i-1}$ by embedding $G_{i-1}$ with $E_{i-1}$ used as initialization rather than directly refining $E_{i}$.

% \cite{deng2019graphzoom} proposed GraphZoom, a multi-level framework of unsupervised graph embedding tasks. The first step of GraphZoom is to  fuses the node attributes and topology of the original graph and construct a new weighted graph. Then they used spectral coarsening to generate the coarsened graphs and generated the embedding on the coarsest graph which has much smaller size. After that, they used graph filters to iteratively refine the graph embeddings to obtain the final result.

\begin{table}[t]
\centering
\begin{small}
\resizebox{\columnwidth}{!}{
\begin{tabular}{|c|c|}
      \hline
      \textbf{Sysmbol}&\textbf{Definition}\\
      \hline
      $G_i$&\ the graph resulting from $i$ iterations of the coarsening process \\
      \hline
      $V_i, E_i$&\ node set and edge set of $G_i$ \\
      \hline
      $O_i, R_i$&\ node types, edge types of $G_i$ \\
      \hline
      $A_i, D_i$&\ the adjacency and degree matrix of $G_i$ \\
      \hline
      $d$&\ dimensionality of the embeddings \\
      \hline
      $m$&\ the total number of coarsening levels \\
      \hline
      $f_e $&\ the base embedding method applicable on $G_i$ \\
      \hline
      $E_i$&\ the embeddings of nodes in $G_i$ \\
      \hline
      $M_{i,i+1}$&\ the matching matrix from $G_i$ to $G_{i+1}$ \\
      
      \hline
\end{tabular}
}
\caption{\label{notation} The table of notations}
\end{small}

\end{table}

% %%%%%%%%%%%%
% %Notation Table Here. 

% % Given a graph $G$ = ($V$,$E$) with a dimensionality $d$ ($d \ll |V|$), we aim to find a vector representation for each node that preserves the properties of the graph. This is done by using a mapping function, $f$, which takes the adjacency matrix of the graph as input and produces a lower-dimensional matrix as output. The goal is to improve the scalability of existing graph embedding methods without compromising on the quality of the resulting embeddings. This is achieved by creating a new and improved method, $f$($\cdot$), that can handle larger datasets while still producing embeddings of equal or better quality than the original method $f$($\cdot$).

% % \subsection{Definition 1  \emph{Graph Embedding}}

\begin{figure}[t]
\centering
\includegraphics[width=.9\linewidth]{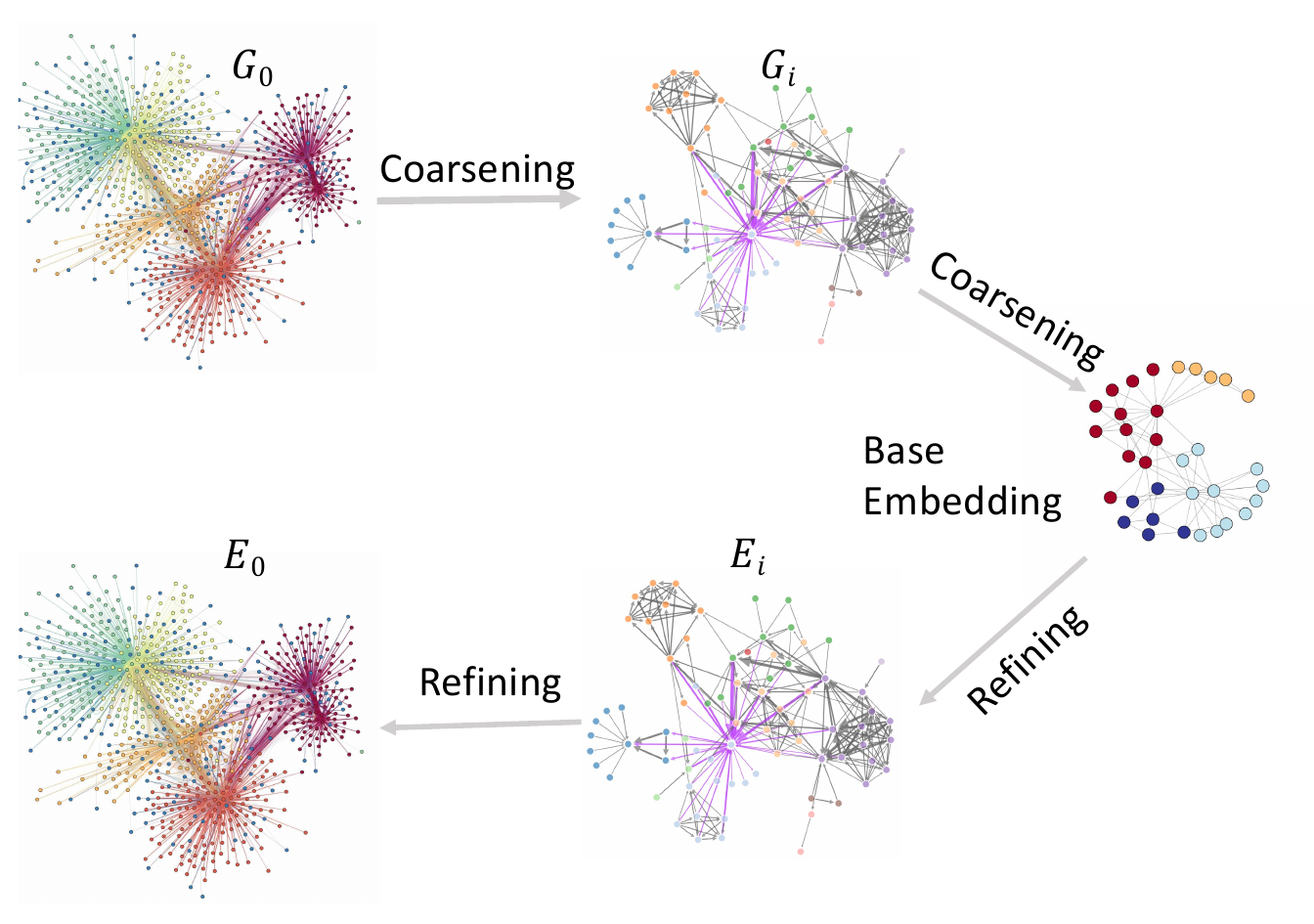}
\caption{Overview of HeteroMILE framework}
\label{overview}
\end{figure}

% % \begin{figure}[t]
% % \centering
% % \includegraphics[scale=0.45]{figures/neighbors.png}
% % \caption{Heterogeneous graph neighbor information}
% % \label{neighbor}
% % \end{figure}

\section{Problem Statement}
We use the notation $G = (V, E)$ to represent a graph, where $V$ is the set of nodes and $E$ is the set of edges in graph $G$. The graph is associated with a node type mapping function $\Phi: V \rightarrow O$ and an edge type mapping function $\Psi: E \rightarrow R$. Here, $O$ represents the set of all node types, and $R$ represents the set of all edge types. It is important to note that for a heterogenous graph, the sum of the cardinalities of $O$ and $R$ must exceed 2, i.e., $|O| + |R| > 2$.
Let $A$ denote the adjacency matrix of the graph. The notations used in this paper are summarized in Table \ref{notation}. Graph embedding can be defined as follows:

\textbf{Scalable Heterogeneous Graph Embedding}
In the context of a heterogeneous graph $G = (V, E)$, where the dimensionality is denoted as $d$ ($d \ll |V|$), our objective is to effectively acquire an embedding model $f: V \rightarrow \mathbb{R}^d$ that successfully captures the semantic aspects of the diverse relationships present within the heterogeneous graph.
% Given a graph $G$ = ($V$,$E$) with a dimensionality $d$ ($d \ll |V|$), we aim to find a vector representation for each node that preserves the properties of the graph. This is done by using a mapping function, $f$, which takes the adjacency matrix of the graph as input and produces a lower-dimensional matrix as output. The goal is to improve the scalability of existing graph embedding methods without compromising on the quality of the resulting embeddings. This is achieved by creating a new and improved method, $f$($\cdot$), that can handle larger datasets while still producing embeddings of equal or better quality than the original method $f$($\cdot$).

% \subsection{Definition 1  \emph{Graph Embedding}}

\begin{figure}[t]
\centering
\includegraphics[width=.85\linewidth]{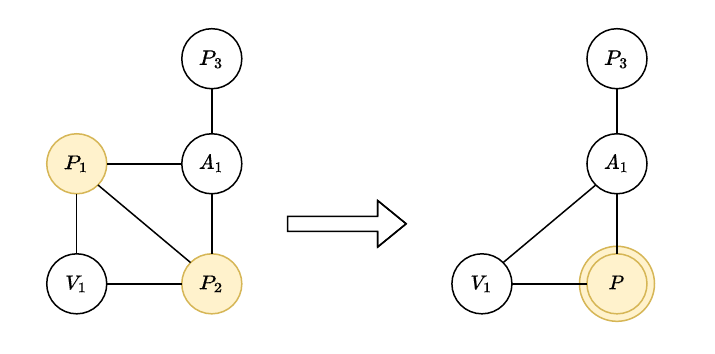}
\caption{Example of matching and merging the nodes}
\label{merge}
\end{figure}

% \begin{figure}[t]
% \centering
% \includegraphics[scale=0.45]{figures/neighbors.png}
% \caption{Heterogeneous graph neighbor information}
% \label{neighbor}
% \end{figure}

\section{Methodology}
The HeteroMILE framework is divided into three stages: graph coarsening, base embedding, and refinement, as illustrated in Figure \ref{overview}. Each stage will be explained in further detail.

\subsection{Graph Coarsening}
During this phase, the initial graph (referred to as $G$ or $G_{0}$) undergoes a repetitive coarsening process, resulting in a sequence of smaller graphs ($G_{1}$, $G_{2}$, ..., $G_{m}$) with a decreasing number of nodes. The coarsening operation involves merging multiple nodes from $G_{i}$ to create larger nodes known as supernodes within $G_{i+1}$. The edges connected to a supernode consist of the combined edges from the original nodes in $G_{i}$. The process of merging nodes to form a supernode is referred to as matching, where nodes of the same type are grouped together. To ensure efficient size reduction of the graph while preserving its overall structure, we propose two matching strategies. Figure \ref{merge} provides an illustrative example of this coarsening process.

\paragraph{\textbf{Jaccard Similarity Matching Strategy: }}
Jaccard Similarity\cite{sathre2022edge} measures the similarity between two sets of data, which is calculated as the size of the intersection divided by the size of the union of two sets. In the context of graphs, Jaccard Similarity can be applied to the sets of 1-hop and 2-hop neighbors of two nodes. By calculating the Jaccard Similarity between each neighbor pair, we merged the nodes in two ways:

\begin{figure}[t]
\centering
\includegraphics[width=.9\linewidth]{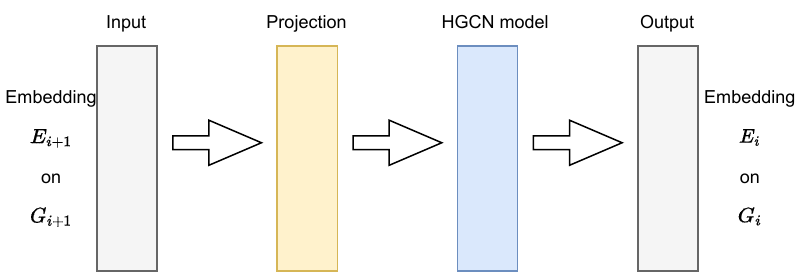}
\caption{Refinement Process of HeteroMILE}
\label{refinement}
\end{figure}

\begin{itemize}
\item{Jaccard Max: } For each unmatched node $u$ in the graph, we find a node $v$ from its 1-hop and 2-hop neighbors (denoted as $N_{(u)}$) which has the maximum Jaccard Similarity. We then collapsed nodes $u$ and $v$ in the same type into one supernode and marked them as matched. This matching process continues until all nodes have been matched or all unmatched nodes do not have unmatched neighbors. In this study, we calculated Jaccard Similarity using the following formula:

\[J(u, v)=\frac{|N_{(u)} \cap N_{(v)} |}{|N_{(u)}\cup N_{(v)} |} \] 

\item{Jaccard weighted random sampling (WRS):} We propose a variant of Jaccard-similarity-based matching by leveraging weighted random sampling (WRS). The difference is that here we randomly pick a neighbor $v$ from $N_{(u)}$ with a probability proportional to the Jaccard similarity instead of selecting the one with maximum similarity. Formally, given a node $u$, we match it with its neighbor $v \in N_{(u)}$ with probability:
\[p(u, v_{j})=\frac{J(u, v_{j})}{\sum_{i \in N_{(u)}}  J(u, v_{i})} \]

% In weighted random sampling (WRS), given an unmatched node $u$, the probability of each pair between $u$ and its neighbor $v_{i}$ from the neighbor sets $N$, denoted as ($u$, $v$), to be selected is determined by its relative Jaccard Similarity weight. WRS can be defined with the following formula:

% \[p(u, v_{j})=\frac{J(u, v_{j})}{\sum_{i \in N}  J(u, v_{i})} \]

\end{itemize}

\paragraph{\textbf{Locality Sensitive Hashing (LSH) Matching Strategy: }}
To further improve the efficiency of graph coarsening, we propose an approximate solution based on locality-sensitive hashing (LSH) \cite{jafari2021survey}.
LSH is an algorithm technique for efficiently finding approximate nearest neighbors in high-dimensional data. It works using hash functions that are designed to produce hash values of those similar nodes. Similar nodes that have common neighbors are highly likely to be grouped together in the same bucket. This greatly reduces the dimensionality of high-dimensional data because the high-dimension input is converted into lower dimensions while preserving the structure, which makes the computation much more efficient.

Formally, for nodes $u$ and $v$ in the same type, we estimate the similarity between $N_{(u)}$ and $N_{(v)}$ as follows: 
\[sim(N_{(u)}, N_{(v)}) = p_{f_{i} \in F}[f_{i}(N_{(u)}) = f_{i}(N_{(v)})] \]
% where $sim$($\cdot$, $\cdot$) $\in$ [0, 1] is a similarity function and $F$ is a family of hash functions. 

In HeteroMILE, a set of $k$ hash functions denoted as $F$, is randomly sampled for the purpose of hashing. Each individual hash function, $f_{i}$ $\in$ $F$, is defined as $f_{i}(u) = \min{\pi(u)}$, where $u \in U$ and $\pi$ represents a permutation. Consequently, for a pair of nodes $u$ and $v$ with a Jaccard similarity of $J(u,v)$, the probability of them having the same hash value can be expressed as $Pr$[$f_{i}(u) = f_{i}(v)$] = $J(u,v)$. To capture and represent this information, a $k$-dimensional vector is assigned as the signature for each node. When two nodes exhibit identical signatures, it indicates that they are likely to possess structural equivalence, which enables their merging. This technique reduces the time complexity from $O(|V|^2)$ to $O(|V|)$.

\paragraph{\textbf{Choice for Coarsening Level: }} 
The coarsening level is a key parameter that can affect both utility and efficiency. % Choosing the right coarsening level in our framework, similar to other frameworks such as graph partitioning and visualization \cite{karypis1998multilevelk, karypis1996parallel, lasalle2013multi}, is based on the specific use case and properties of the graph. 
Later we empirically show that using a small number of coarsening levels (typically between 2 and 4) results in high-quality embeddings with a good balance of speedup for medium-sized graphs (with less than 1,000,000 nodes). For larger graphs, the embeddings maintain their high quality even with higher coarsening levels (i.e., between 4 and 6), which leads to an even greater speedup. % We will provide more information about this in the experimental results section.

\subsection{Base Embedding}
The coarsening process significantly decreased the size of the graph, with the potential to halve it at each iteration. We perform this process for a fixed number of iterations $m$ and then use a graph embedding technique $f_e$ on the final coarsest graph $G_{m}$. The embeddings produced in $G_{m}$ are referred to as $E_{m}$. Because the graph size is dramatically reduced, the computational time decreases dramatically. 

\begin{table}[t]
\centering
\begin{small}
\resizebox{\columnwidth}{!}{
\begin{tabular}{|c|c|c|}
      \hline
      \textbf{Datasets}&\textbf{Nodes}&\textbf{Edges}\\
      \hline
      AcademicII&\makecell{\# author (A): 28,646 \\ \# paper (P): 21,044\\ \# venue (V): 18}&\makecell{\# A-P: 69,311 \\ \# P-P: 46,931 \\ \# P-V: 21,044} \\
      \hline
      DBLP&\makecell{\# author (A): 4,057 \\ \# paper (P): 14,328 \\ \# term (T): 7,723 \\ \# venue (V): 20 }&\makecell{\# A-P: 19,645 \\ \# P-T: 85,810 \\ \# P-V: 14,328 }\\
      \hline
      IMDB&\makecell{\# movie (M): 4,278 \\ \# director (D): 2,081 \\ \# actor (A): 5,257}&\makecell{\# M-D: 4,278 \\ \# M-A: 12,828}\\
      \hline
      OGB\_MAG&\makecell{\# paper (P): 736,389 \\ \# author (A):  1,134,649 \\ \# institution (I): 8,740 \\ \# field (F): 59,965}&\makecell{\# P-A: 7,145,660 \\ \# P-F: 7,505,078 \\ \# P-P: 5,416,271 \\ \# A-I: 1,043,998 }\\
      \hline
      % &1,939,473&21,111,007
\end{tabular}
}
\end{small}
\caption{\label{dataset} Dataset Information}
\end{table}

\begin{algorithm}[t] 
\caption{HeteroMILE Algorithm for Graph Embedding}\label{pipeline_algo}
    \begin{flushleft}
        \textbf{Input:} Given an input graph denoted as $G_0 = (V_0, E_0)$, a specified number of coarsening levels $m$, and a base embedding method represented as $f_e$. \\
        \textbf{Output:} Graph embeddings, denoted as $E_0$, generated on graph $G_0$.
    \end{flushleft}
    \begin{algorithmic}[1]
%    \Procedure{MyProcedure}{$x,y$}
%     % Input:
%     \Comment{Input: x}
%     % Output:
%     \Comment{Output:y}
    \State The input graph $G_0$ is coarsened into a sequence of coarsened graphs $G_1, G_2, ..., G_m$ using either the Jaccard Similarity matching or LSH matching method
    \State Apply the base embedding method $f_e$ on the coarsest graph $G_m$ to obtain the graph embeddings $E_m$.
    \State Learn the weights $W$ by optimizing the loss function.

    \For {$ i = (m-1) ... 0$}
    \State Compute the projected embeddings $E_i^p$ for each graph $G_i$.
    \State Compute the refined embeddings $E_i$ using refinement models equations.
    \EndFor
    \State Return the graph embeddings $E_0$ on $G_0$.
    \end{algorithmic}
    
\end{algorithm}

\subsection{Refinement}

The HeteroMILE framework's final stage focuses on refining the embeddings to derive the node embeddings of the original graph $G_{0}$ from the coarsened graph $G_{m}$. We start by addressing the simpler task of inferring the embeddings $E_{i}$ for a graph $G_{i}$ using its coarsened version $G_{i+1}$, the node embeddings $E_{i+1}$ from $G_{i+1}$ and the matching matrix $M_{i, i+1}$ as shown in Figure \ref{refinement}. Once this step is completed, we can systematically apply this approach to consecutive pairs of graphs, starting with $G_{m}$ and progressing towards $G_{0}$. As a result, we gradually obtain the node embeddings for $G_{0}$. To accomplish this, we utilize a Heterogeneous Graph Convolutional Network model, which aids in refining the embeddings and improving their quality.

\paragraph{\textbf{Heterogeneous Graph Convolutional Network for Refinement Learning: }}

By leveraging the matching information between two consecutive graphs, specifically $G_{i}$ and $G_{i+1}$, we can utilize a projection technique to transfer the node embeddings from the coarser graph $G_{i+1}$ to the finer graph $G_{i}$. This process involves mapping the embeddings from $G_{i+1}$ to the corresponding nodes in $G_{i}$.

\[E_{i}^{p}=M_{i,i+1}E_{i+1} \]

In this case, the embedding of a supernode is duplicated across its corresponding original node(s). The duplicated embeddings, denoted as $E_{i}^{p}$, correspond to the projected embeddings from $G_{i+1}$ to $G_{i}$, or simply referred to as projected embeddings. These embeddings capture the structure information transferred from the coarser graph to the finer graph and serve as a representation of the nodes in the refined embedding space. The simple projection method preserves some information from the node embeddings; however, it has a clear drawback in that if nodes are grouped and collapsed into a supernode during the coarsening stage, they will have identical embeddings. This issue becomes increasingly severe as the embedding refinement process is repeated multiple times, starting from $G_{m}$ and working down to $G_{0}$. 
To address this constraint, we introduce a refinement model that leverages the capabilities of a Heterogeneous Graph Convolutional Network (HGCN) \cite{yang2021interpretable}. The proposed model enables the refinement of the embeddings by taking advantage of the projected embeddings $E_{i}^{p}$, which are derived from the initial embedding method, and the adjacency matrix $A_{i}$, which is obtained from the input graph. These components serve as input to the refinement process, enhancing the quality and accuracy of the embeddings. Using these inputs, the refinement model generates embeddings $E_{i}$ for the graph $G_{i}$.

% To overcome this limitation, we propose to use a Heterogeneous Graph Convolutional Network (HGCN) for refining the embeddings (as proposed \cite{yang2021interpretable}). Specifically, we propose a graph neural network model, $E_{i} = R(E_{i}^{p}, A_{i})$, that generates embeddings $E_{i}$ for the graph $G_{i}$, by utilizing the projected embeddings $E_{i}^{p}$ (from the initial method) and the adjacency matrix $A_{i}$ (from the input graph) as inputs.

Consider $O_N$ refers to a neighbor node type of $O_i$. $V^O_i$ as the set comprising nodes of type $O_i$, while $N_{O_i}$ represents the set of all neighbor node types of $O_i$ (i.e. the set comprising nodes of type $O_N$). The relationship linking $O_N$ and $O_i$ is denoted as $<O_N, O_i>$ or $O_N \rightarrow O_i$.

When considering the graph $G$, we take into account the expedited approximation of heterogeneous graph convolution proposed in \cite{yang2021interpretable}. In this neural network model, the subsequent layer is
\[H_{i}^{O_i'}=\sigma(a_{i}^{Self-O_i}\cdot Z_{i,:}^{Self-O_i} + \sum_{O_N \in N_{O_i}} a_{i}^{O_N \rightarrow O_i} \cdot Z_{i,:}^{O_N \rightarrow O_i}) \]

Here, $\sigma$ represents the nonlinearity, and the subscript $i (i,:)$ refers to the $i$-th node in $V^{O_i}$. The terms $a_{i}^{Self-O_i}$ and $a_{i}^{O_N \rightarrow O_i}$ denote the normalized attention coefficients for $V^{O_i}$ and $V^{O_N}$, respectively. $Z^{Self-O_i}$ denotes the projected representation of $V^{O_i}$, while $Z^{O_N \rightarrow O_i}$ represents the aggregated representations from $V^{O_N}$ to $V^{O_i}$. The projection is computed as follows:

\[Z^{Self-O_i}=H^{O_i} \cdot W^{Self-O_i} \]
\[Z^{O_N \rightarrow O_i} = \hat{A}^{O_i - O_N} \cdot H^{O_N} \cdot W^{O_N \rightarrow O_i}, O_N \in N_{O_i}\]

In this context, $H^O_i$ refers to the hidden representation of $V^O_i$ in the preceding layer, $\hat{A}$ indicates the adjacency matrix that has been row-normalized, and $W$ denotes a weight matrix. In the initial scenario, specifically when layer $n$ = 1, $H^{O_i}[1]$ is assigned as $E^{O_i}$, representing the node embedding.

% \textbf{add a calculation flow here. }
\begin{figure*}[!t]
\centering
\includegraphics[width=.8\linewidth]{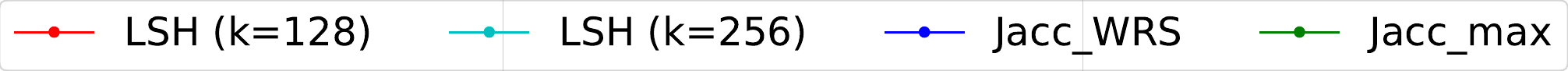}\\
\subfloat[AcademicII (Micro-F1)]{\includegraphics[width=0.247\linewidth]{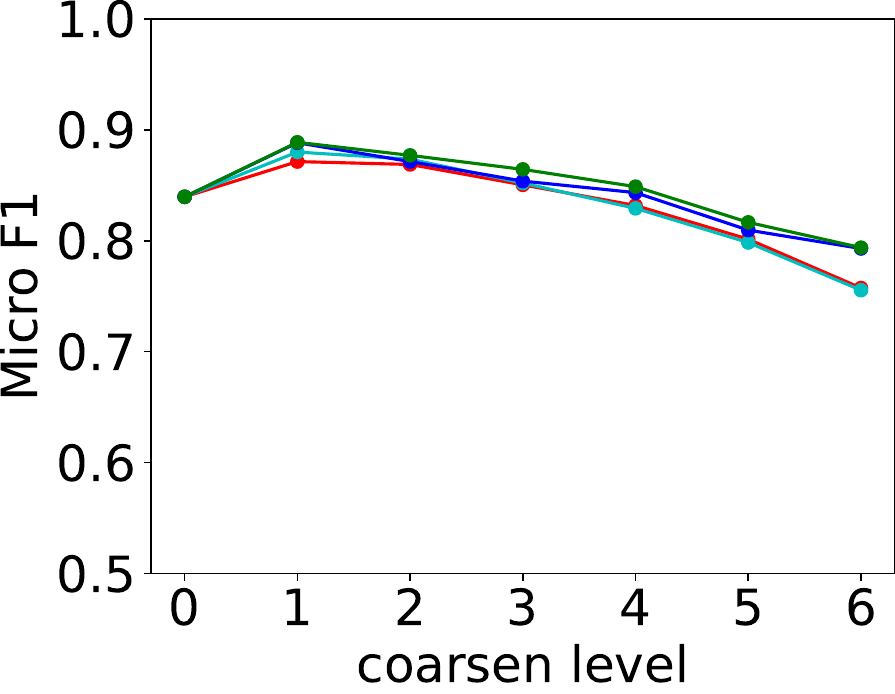}
\label{fig:academicII_f1}}
\subfloat[DBLP (Micro-F1)]{\includegraphics[width=0.247\linewidth]{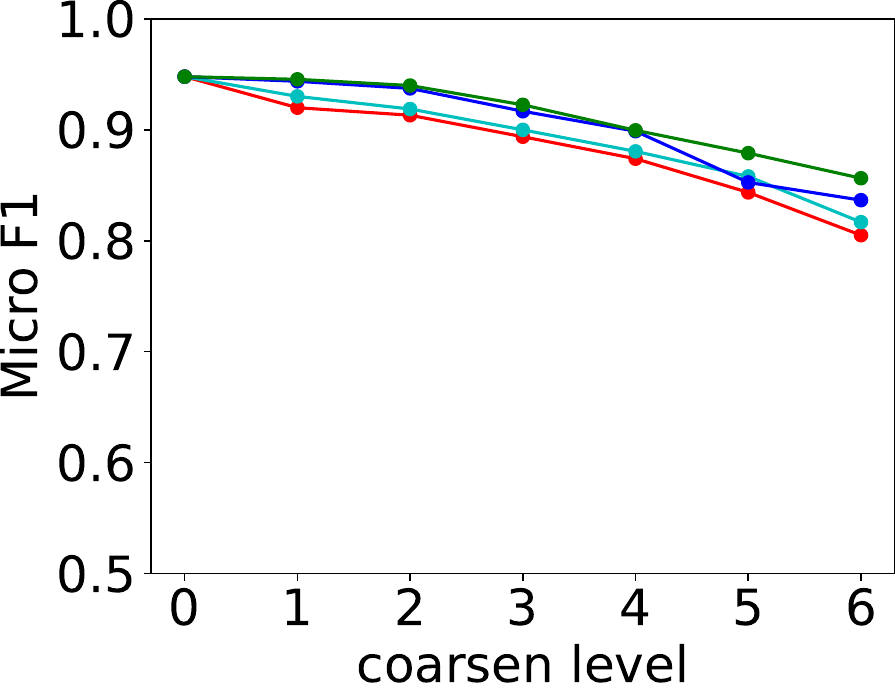}
\label{fig:dblp_f1}}
\subfloat[IMDB (Micro-F1)]{\includegraphics[width=0.247\linewidth]{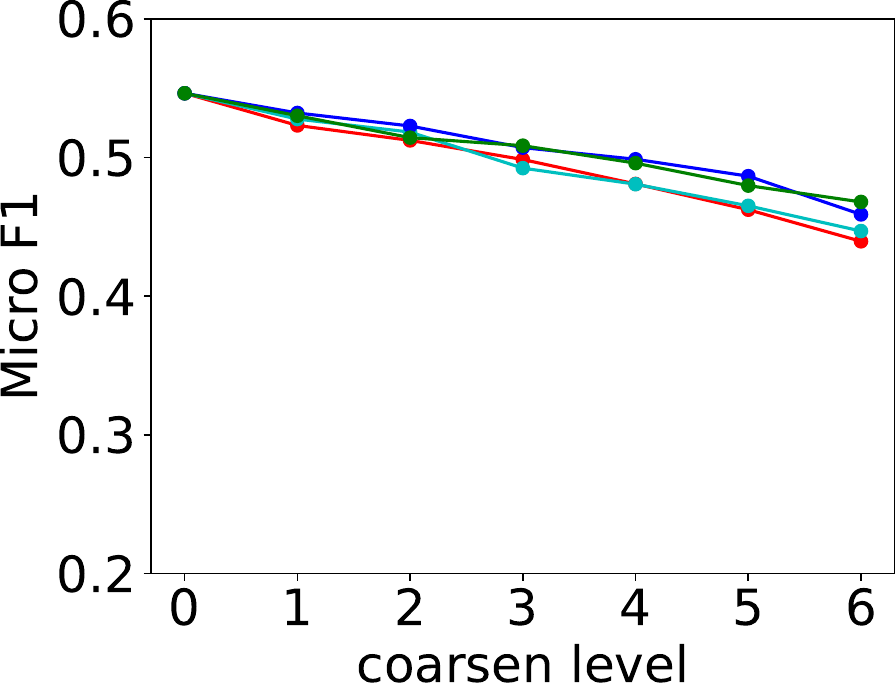}
\label{fig:imdb_f1}}
\subfloat[OGB\_MAG (Micro-F1)]{\includegraphics[width=0.25\linewidth]{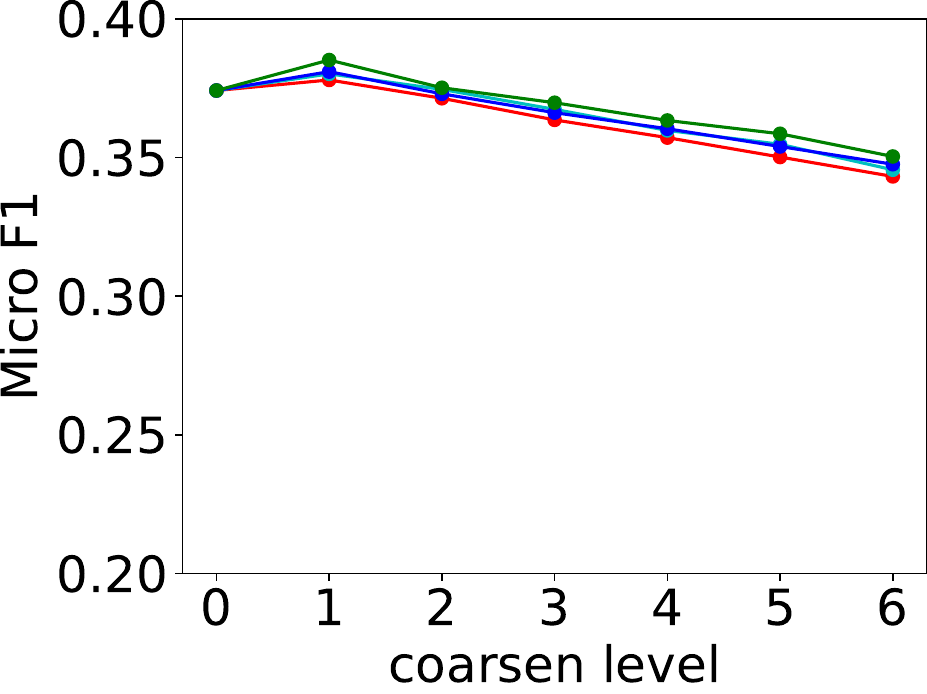}
\label{fig:ogb_f1}} \\

\subfloat[AcademicII (AUROC)]{\includegraphics[width=0.247\linewidth]{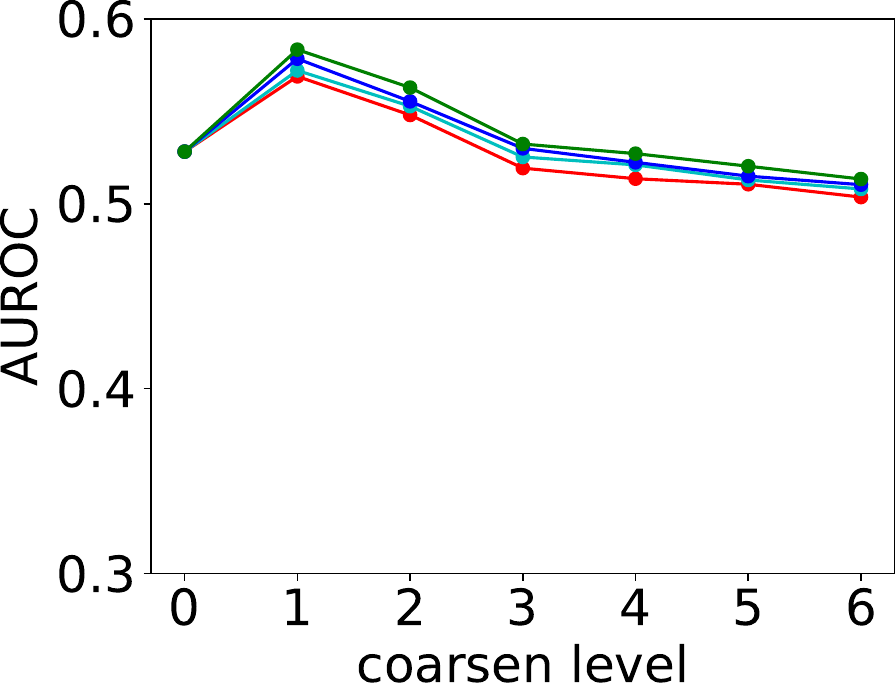}
\label{fig:academicII_auroc}}
\subfloat[DBLP (AUROC)]{\includegraphics[width=0.247\linewidth]{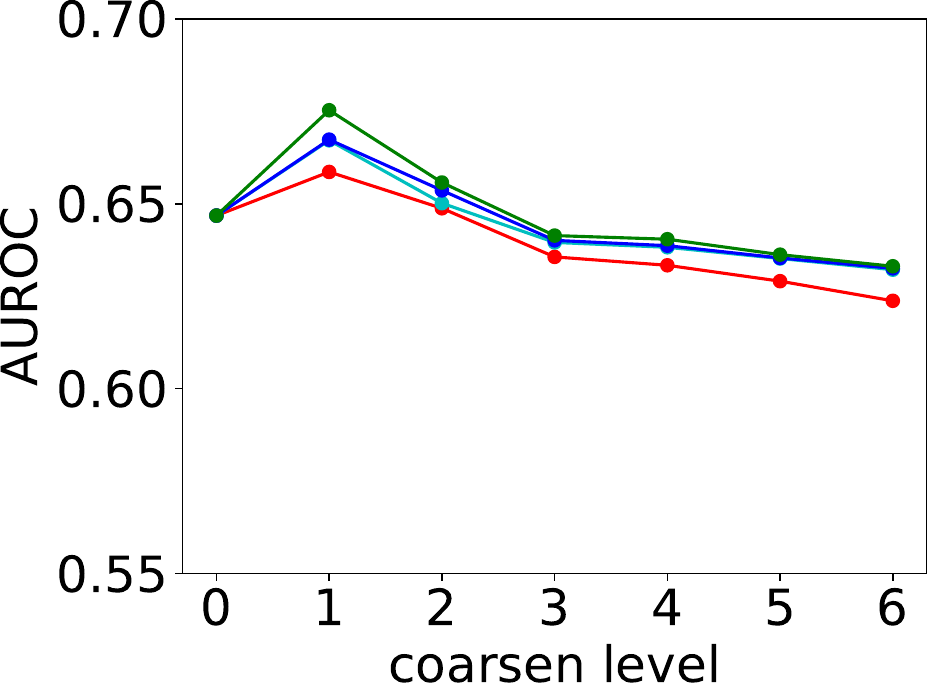}
\label{fig:dblp_auroc}}
\subfloat[IMDB (AUROC)]{\includegraphics[width=0.247\linewidth]{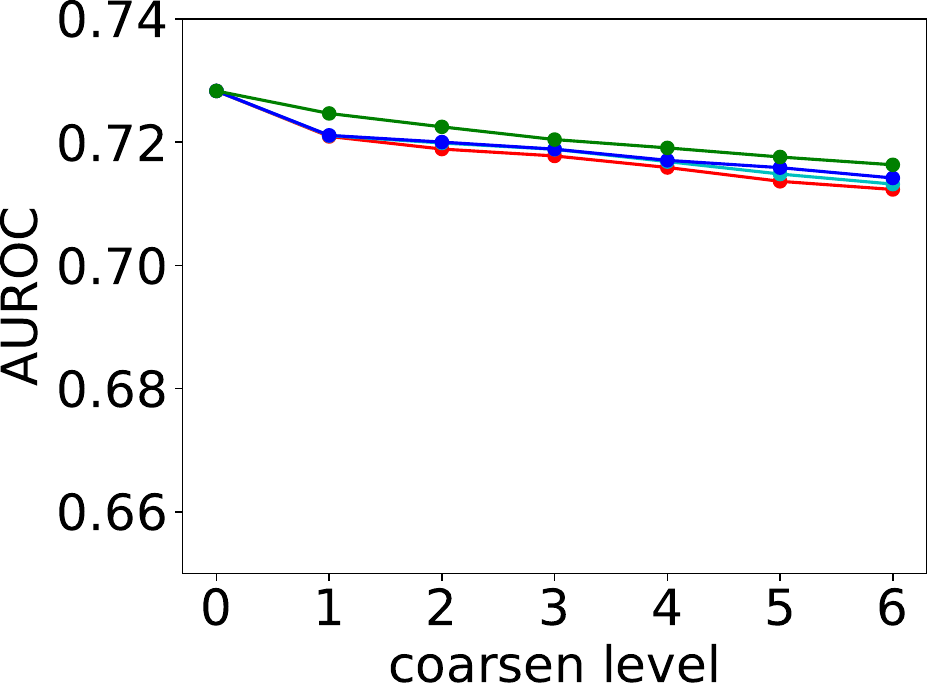}
\label{fig:imdb_auroc}}
\subfloat[OGB\_MAG (AUROC)]{\includegraphics[width=0.253\linewidth]{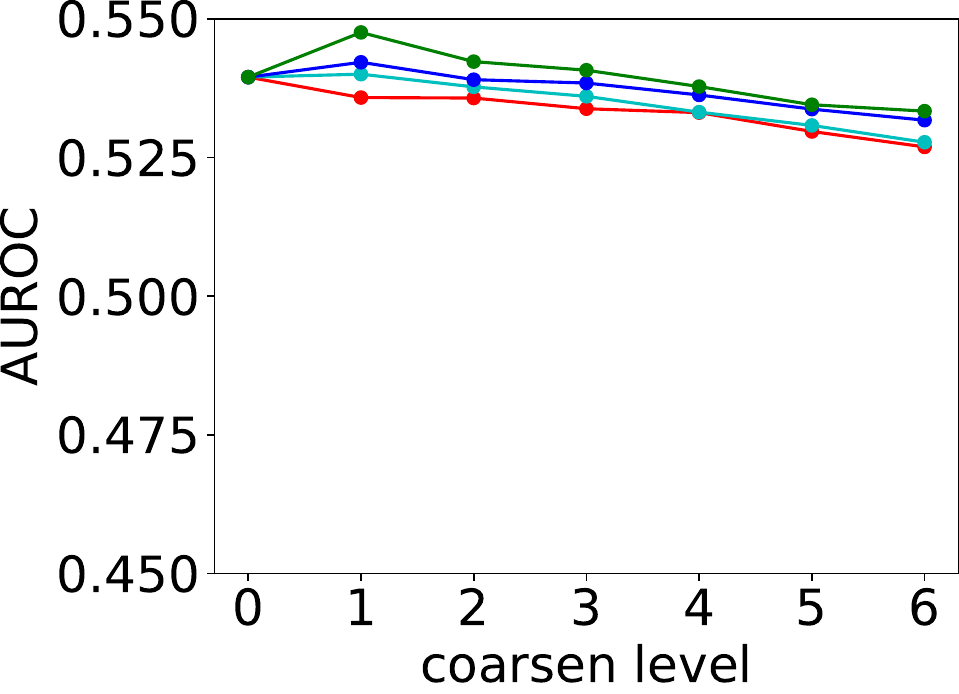}
\label{fig:ogb_auroc}} \\

\subfloat[AcademicII (Time)]{\includegraphics[width=0.247\linewidth]{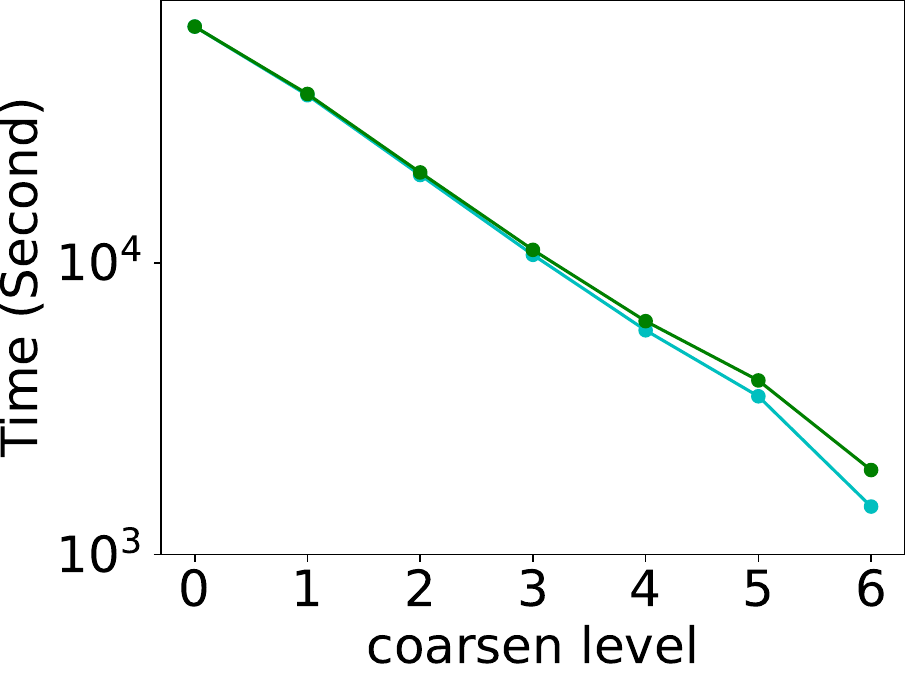}
\label{fig:academicII_time}}
\subfloat[DBLP (Time)]{\includegraphics[width=0.247\linewidth]{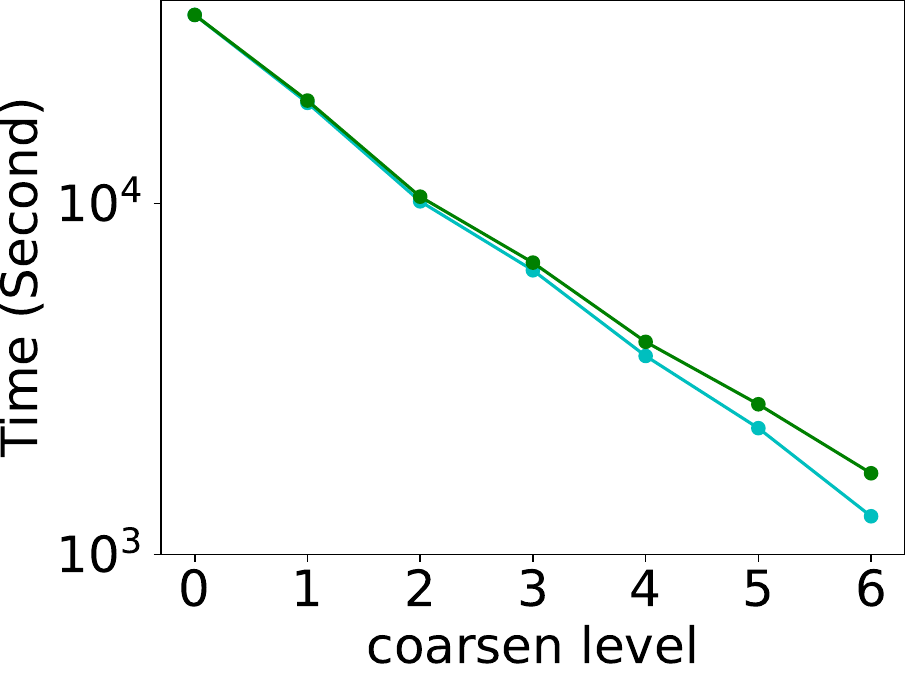}
\label{fig:dblp_time}}
\subfloat[IMDB (Time)]{\includegraphics[width=0.247\linewidth]{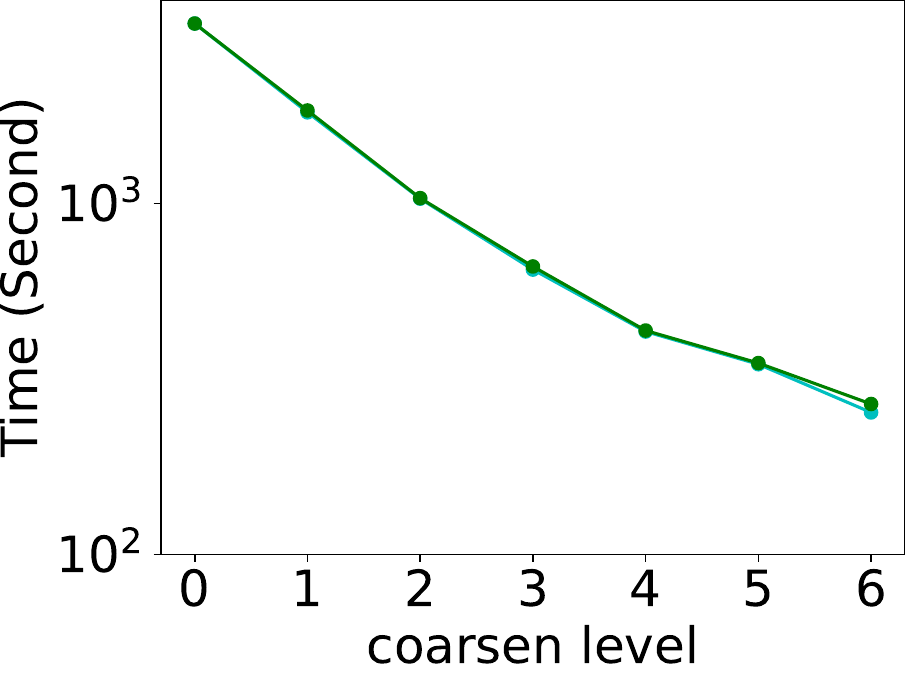}
\label{fig:time_time}}
\subfloat[OGB\_MAG (Time)]{\includegraphics[width=0.247\linewidth]{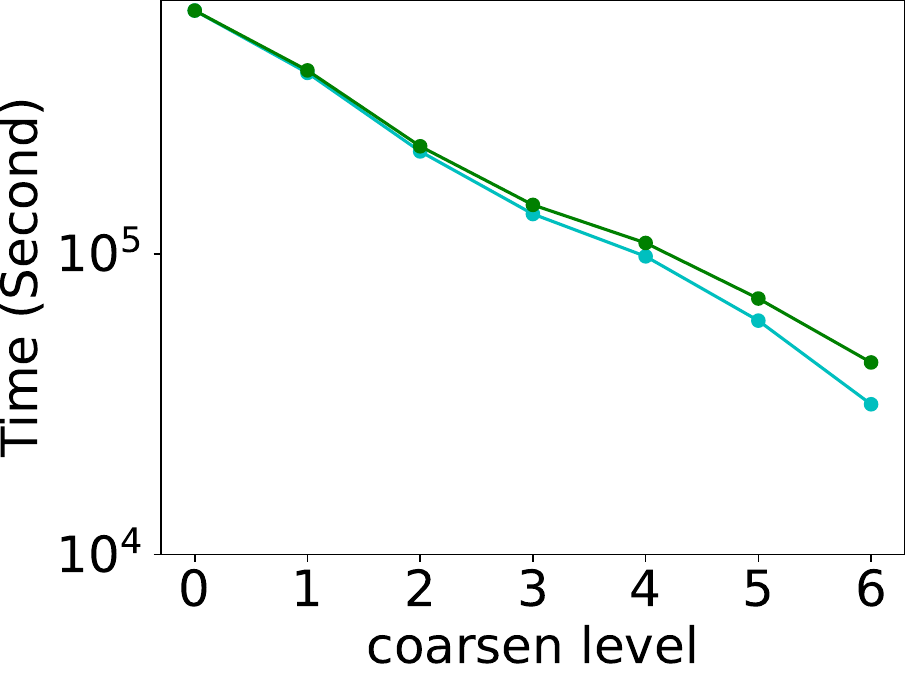}
\label{fig:ogb_time}}
\caption{The performance of HeteroMILE using metapath2vec as the base embedding method varies as the number of coarsening levels increases, as depicted by the color scheme. The results for node classification, measured by the Micro-F1 score, and link prediction, measured by AUROC, are presented in the first and second rows, respectively. The running time, displayed in the third row, is plotted on a logarithmic scale. Notably, the running time lines of Jacc\_WRS and Jacc\_max overlap, similar to LSH (k=128) and LSH (k=256). "level = 0" represents the original embedding method without HeteroMILE.}
\label{performance}
\end{figure*}

\begin{figure*}[!t]
\centering
\includegraphics[width=.8\linewidth]{figures/legend_preliminary.pdf}\\
\subfloat[AcademicII (Micro-F1)]{\includegraphics[width=0.247\linewidth]{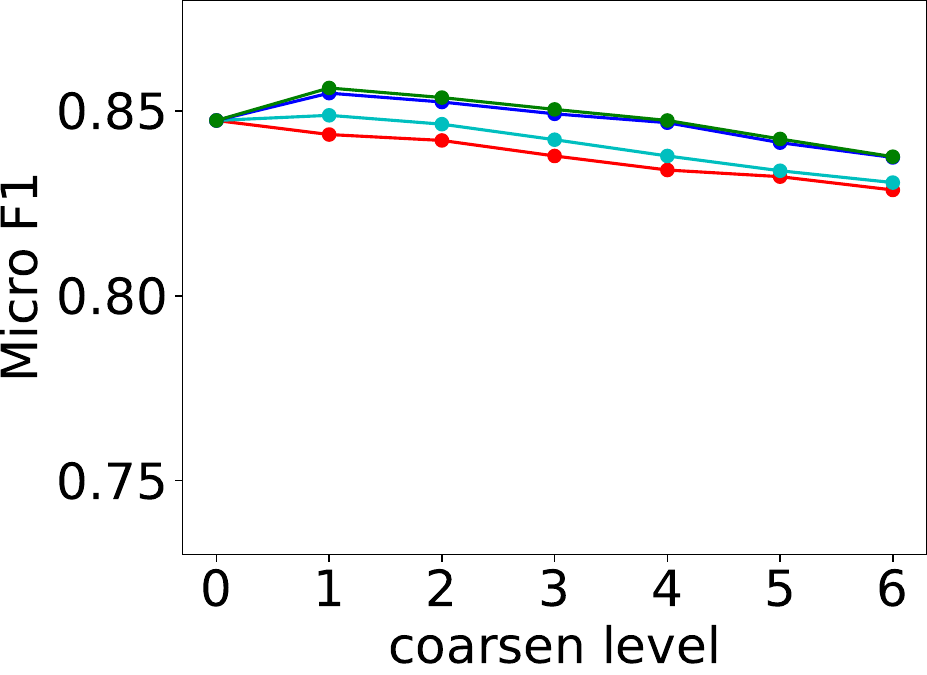}
\label{fig:g_academicII_f1}}
\subfloat[DBLP (Micro-F1)]{\includegraphics[width=0.247\linewidth]{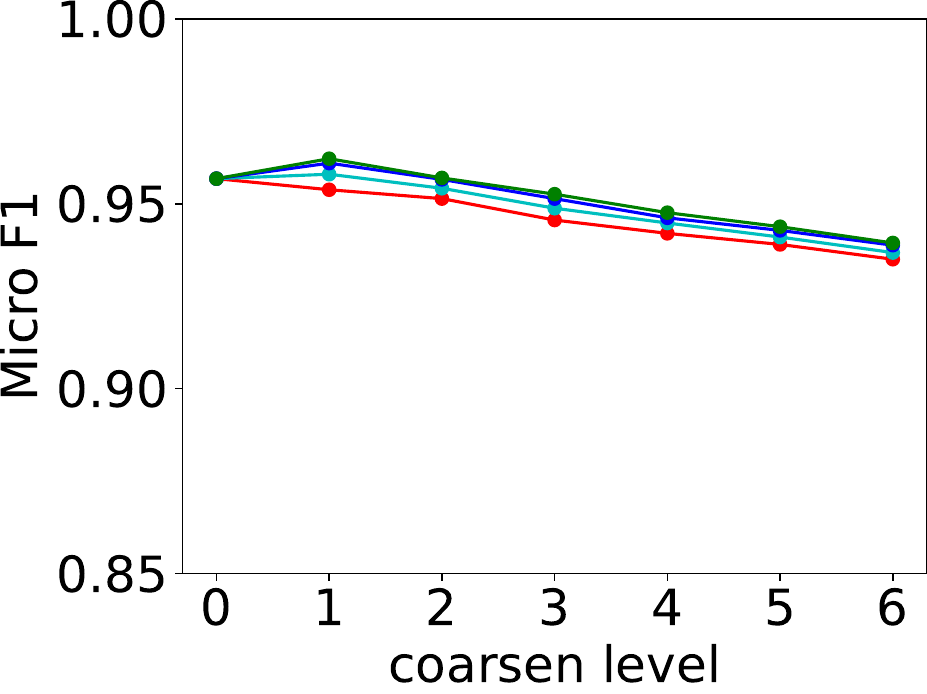}
\label{fig:g_dblp_f1}}
\subfloat[IMDB (Micro-F1)]{\includegraphics[width=0.247\linewidth]{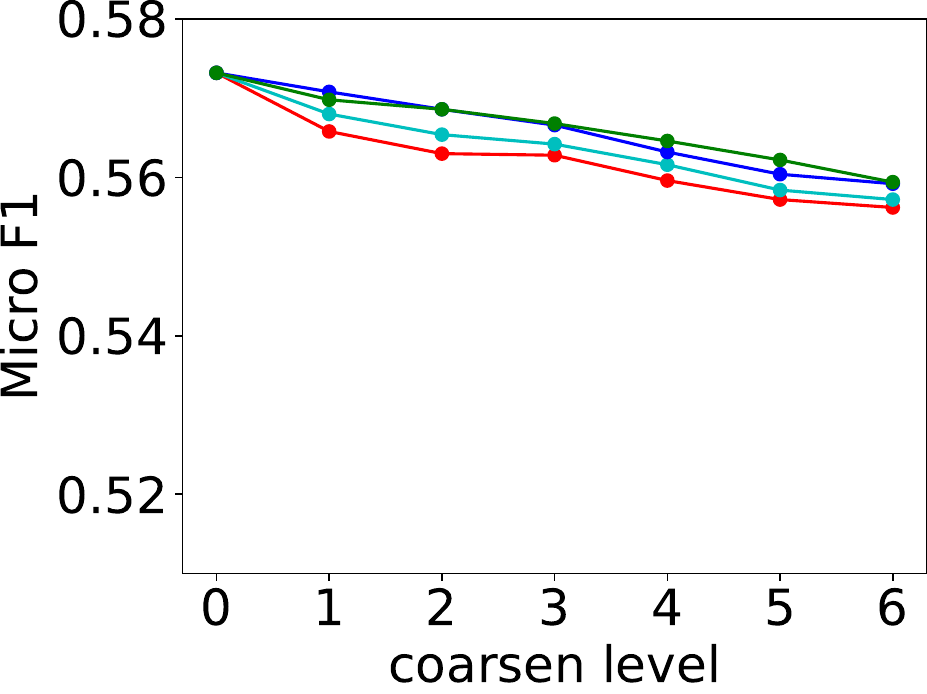}
\label{fig:g_imdb_f1}}
\subfloat[OGB\_MAG (Micro-F1)]{\includegraphics[width=0.25\linewidth]{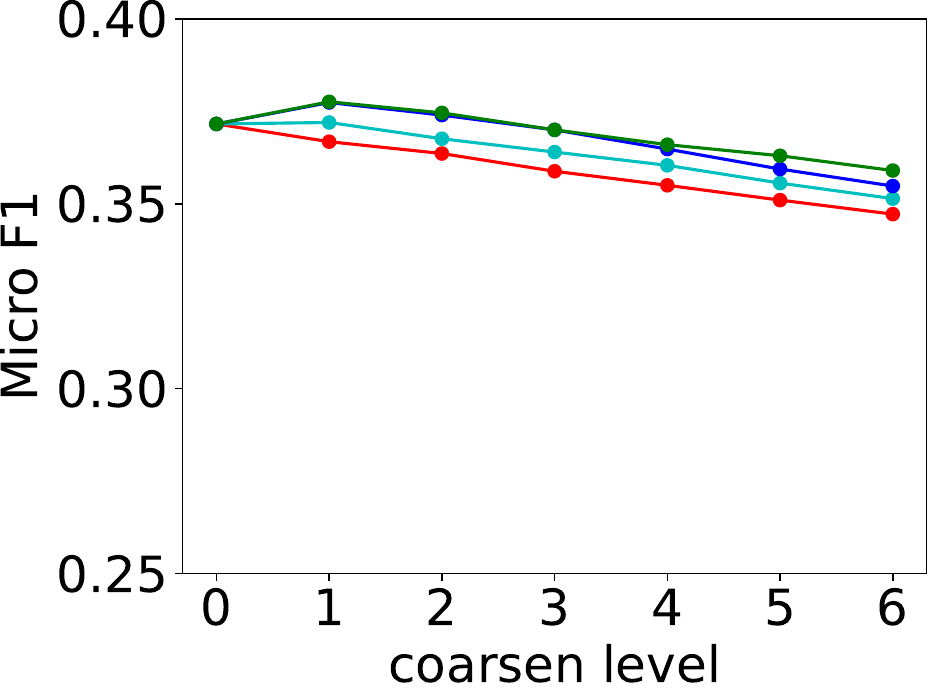}
\label{fig:g_ogb_f1}} \\

\subfloat[AcademicII (AUROC)]{\includegraphics[width=0.247\linewidth]{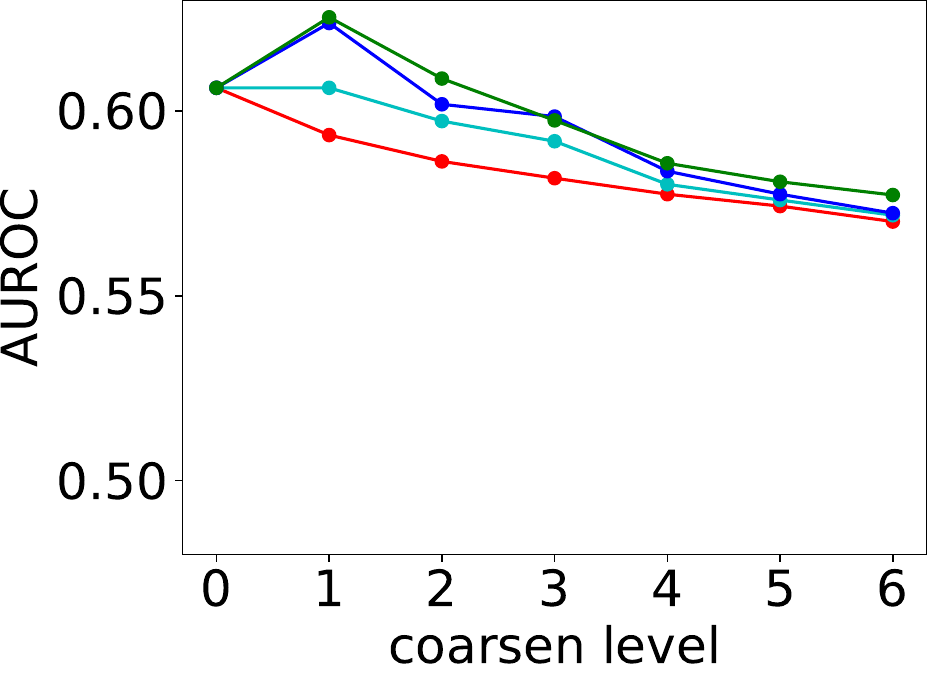}
\label{fig:g_academicII_auroc}}
\subfloat[DBLP (AUROC)]{\includegraphics[width=0.247\linewidth]{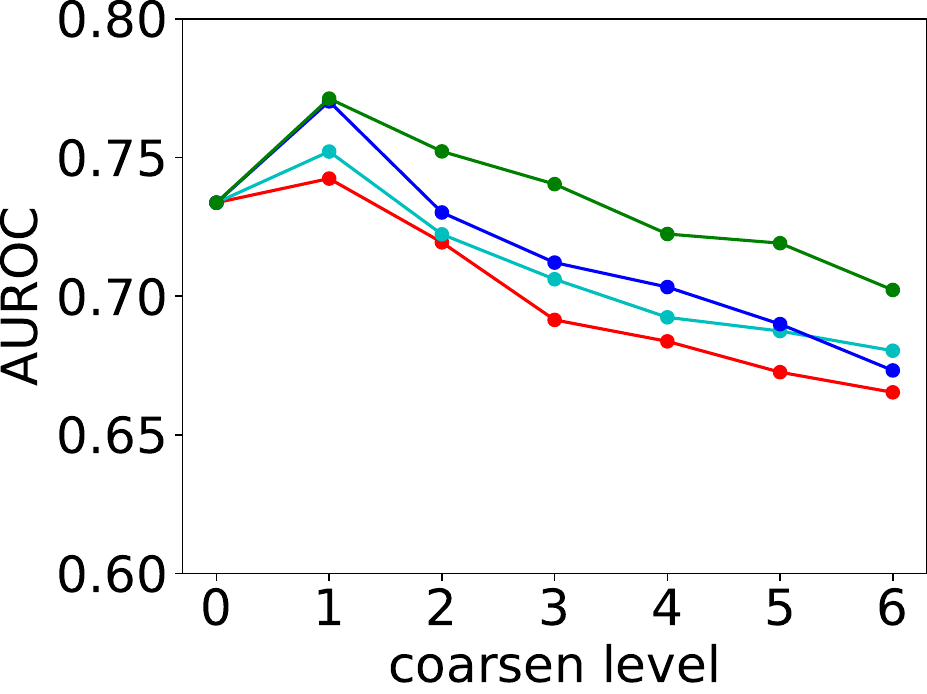}
\label{fig:g_dblp_auroc}}
\subfloat[IMDB (AUROC)]{\includegraphics[width=0.247\linewidth]{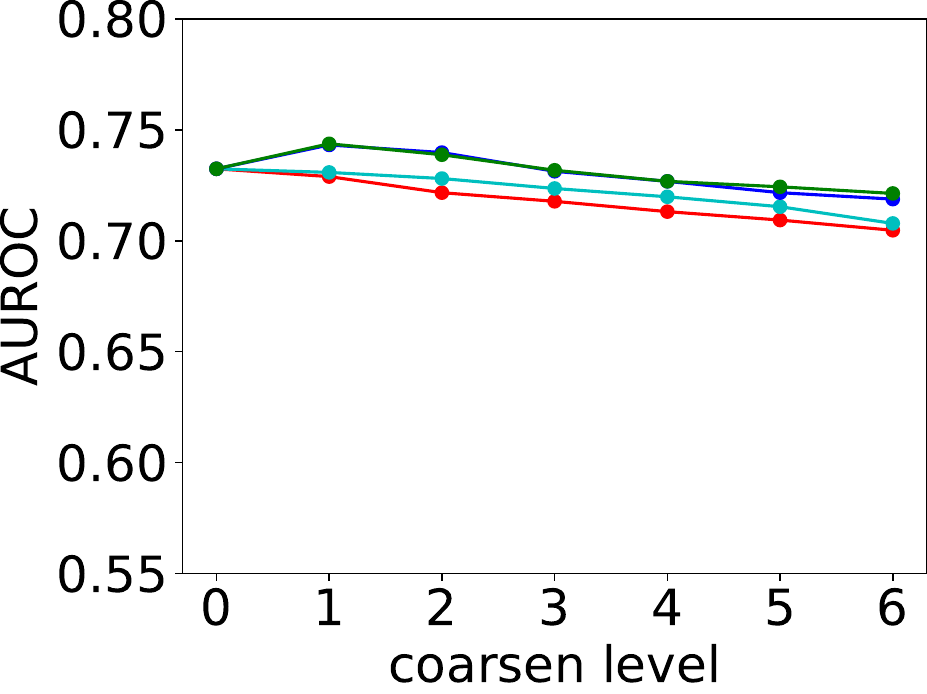}
\label{fig:g_imdb_auroc}}
\subfloat[OGB\_MAG (AUROC)]{\includegraphics[width=0.253\linewidth]{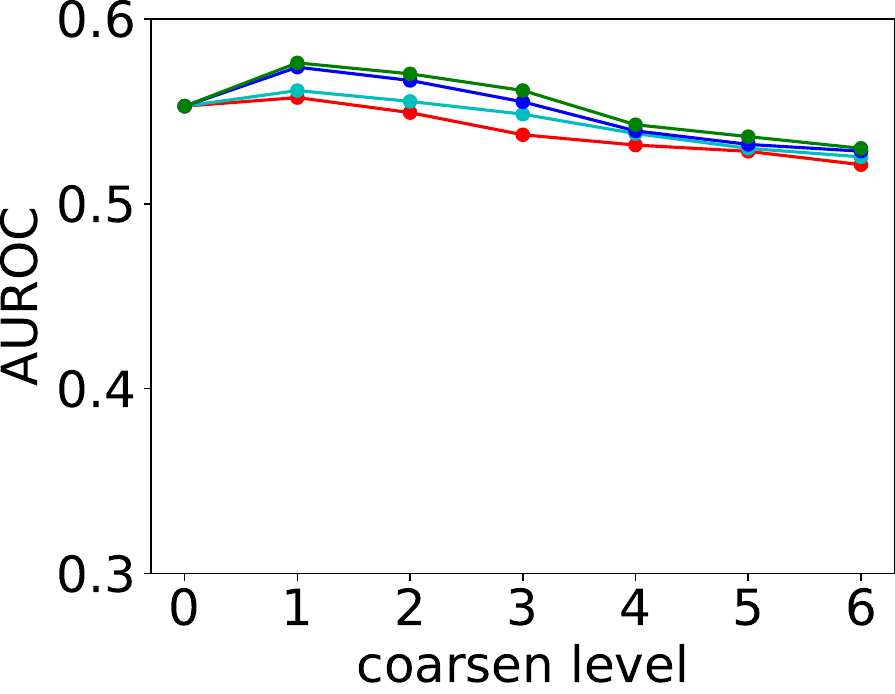}
\label{fig:g_ogb_auroc}} \\

\subfloat[AcademicII (Time)]{\includegraphics[width=0.247\linewidth]{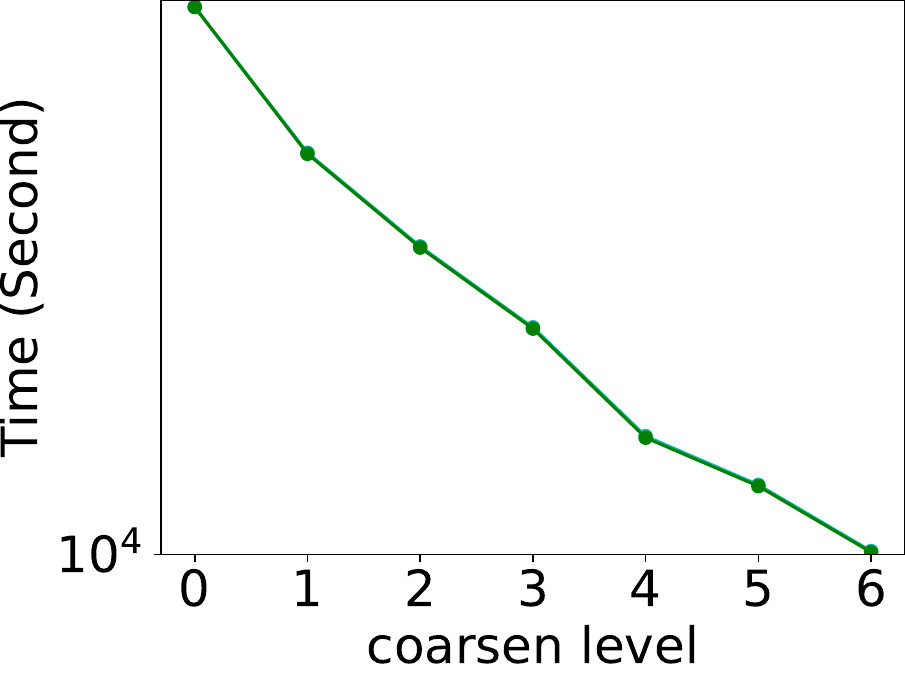}
\label{fig:g_academicII_time}}
\subfloat[DBLP (Time)]{\includegraphics[width=0.247\linewidth]{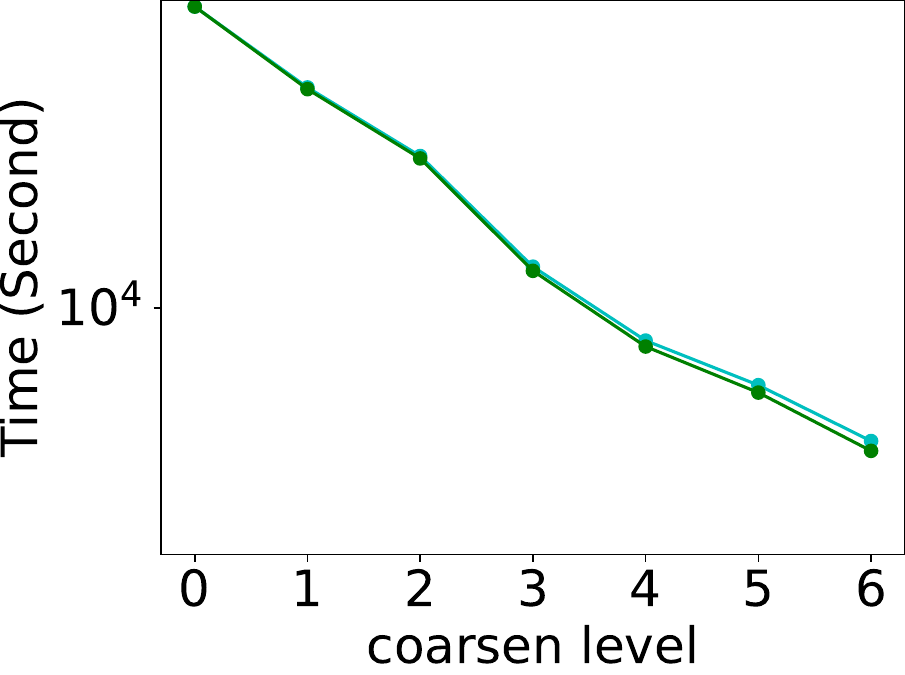}
\label{fig:g_dblp_time}}
\subfloat[IMDB (Time)]{\includegraphics[width=0.247\linewidth]{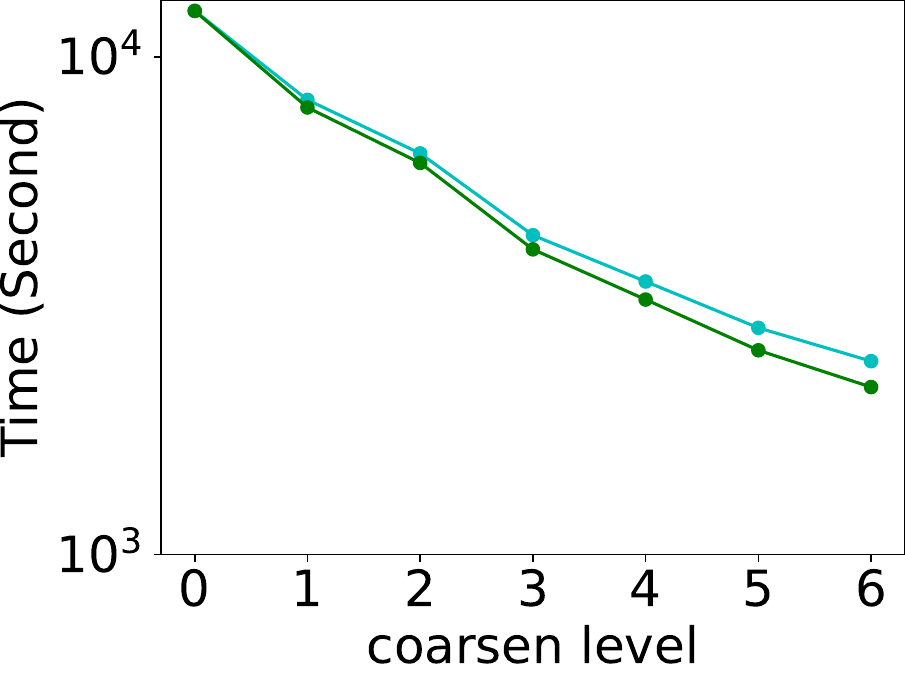}
\label{fig:g_imdb_time}}
\subfloat[OGB\_MAG (Time)]{\includegraphics[width=0.247\linewidth]{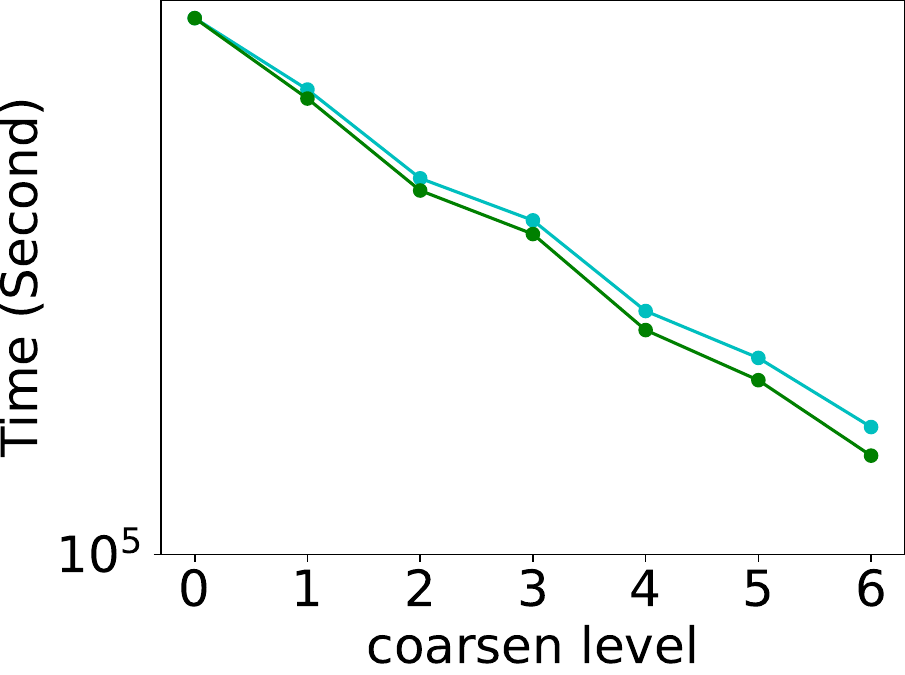}
\label{fig:g_ogb_time}}
\caption{
The performance of HeteroMILE using GATNE as the base embedding method varies as the number of coarsening levels increases, as depicted by the color scheme. The results for node classification, measured by the Micro-F1 score, and link prediction, measured by AUROC, are presented in the first and second rows, respectively. The running time, displayed in the third row, is plotted on a logarithmic scale. Notably, the running time lines of Jacc\_WRS and Jacc\_max overlap, similar to LSH (k=128) and LSH (k=256). "level = 0" represents the original embedding method without HeteroMILE.}
\label{performance_gatne}
\end{figure*}

\paragraph{\textbf{Loss Function:}}
The aim of the refinement model is to acquire the weights $W$ for each layer and different node types through the utilization of a heterogeneous graph convolution model. The objective of this model is to generate predictions for the embeddings $E_{i}$ of a given graph $G_{i}$. The "ground-truth" embeddings are obtained by applying a projected embedding $E_i^p$, and the loss function is formulated as the mean square error between the predicted embeddings and the ground-truth embeddings. 
In our approach, the weights $W$ are initially learned on the coarsest graph and then shared across all levels during the refinement process. This sharing of weights allows for efficient learning and promotes consistency in the embeddings across different levels. The loss function used in our approach is the mean square error, which quantifies the discrepancy between the predicted embeddings and the ground-truth embeddings. The loss is defined as follow:

\[L = \frac{1}{V_m} |E_m - H|^2 \]
The detailed pipeline for HeteroMILE is presented in Algorithm \ref{pipeline_algo}.

\begin{table*}[t]
\centering
\begin{tabular}{|c|c|c|c|c|}
      \hline
      \textbf{Datasets}&\textbf{Coarsening Method}&\textbf{F1}&\textbf{AUROC}&\textbf{Coarsening Time}\\
      \hline
      AcademicII&\makecell{Jacc\_WRS \\ Jacc\_MAX \\ LSH(k=128) \\ LSH(k=256)}&\makecell{ 0.8482 \\ 0.8503 \\ 0.8378 \\ 0.8422}&\makecell{ 0.5984\\ 0.5974 \\ 0.5817  \\ 0.5917 }&\makecell{464.21078 \\ 465.16261 \\ 38.884464 \\ 39.323140}\\
      \hline
      DBLP&\makecell{Jacc\_WRS \\ Jacc\_MAX \\ LSH(k=128) \\ LSH(k=256)}&\makecell{ 0.9514 \\ 0.9526 \\ 0.9456 \\ 0.9488}&\makecell{ 0.7121\\ 0.7404 \\ 0.6914   \\ 0.7061}&\makecell{355.23242 \\ 356.32588 \\ 22.213039 \\ 23.034396}\\
      \hline
      IMDB&\makecell{Jacc\_WRS \\ Jacc\_MAX \\ LSH(k=128) \\ LSH(k=256)}&\makecell{ 0.5666\\ 0.5668  \\ 0.5628 \\ 0.5642}&\makecell{ 0.7313\\ 0.7317 \\ 0.7177 \\ 0.7235}&\makecell{9.2708401 \\ 9.2509674 \\ 2.6778357 \\ 2.7453285}\\
      \hline
      OGB\_MAG&\makecell{Jacc\_WRS \\ Jacc\_MAX \\ LSH(k=128) \\ LSH(k=256)}&\makecell{0.3700 \\ 0.3701 \\ 0.3588 \\ 0.3640}&\makecell{0.5551 \\ 0.5612 \\ 0.5373 \\ 0.5484}&\makecell{11235.346 \\ 11214.024 \\ 1338.1838 \\ 1379.7852}\\
      \hline
\end{tabular}
\caption{\label{coarsen_compare} The performance and coarsening time in seconds using different coarsening strategies}
\end{table*}

\begin{figure*}[!t]
\centering
\includegraphics[width=.4\linewidth]
{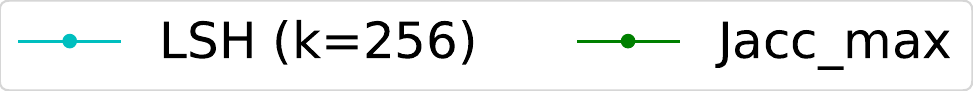}\\

%\subfloat[AcademicII (Micro-F1)]{\includegraphics[width=0.247\linewidth]{figures/academic_comp.pdf}
%\label{fig:academicII_comp_f1}}
%\subfloat[DBLP (Micro-F1)]{\includegraphics[width=0.247\linewidth]{figures/dblp_comp.pdf}
%\label{fig:dblp_comp_f1}}
%\subfloat[IMDB (Micro-F1)]%{\includegraphics[width=0.247\linewidth]{figures/imdb_comp.pdf}
%\label{fig:imdb_comp_f1}}
%\subfloat[OGB\_MAG (Micro-F1)]{\includegraphics[width=0.25\linewidth]{figures/ogb_comp.pdf}
%\label{fig:ogb_comp_f1}} \\

\subfloat[AcademicII]{\includegraphics[width=0.247\linewidth]{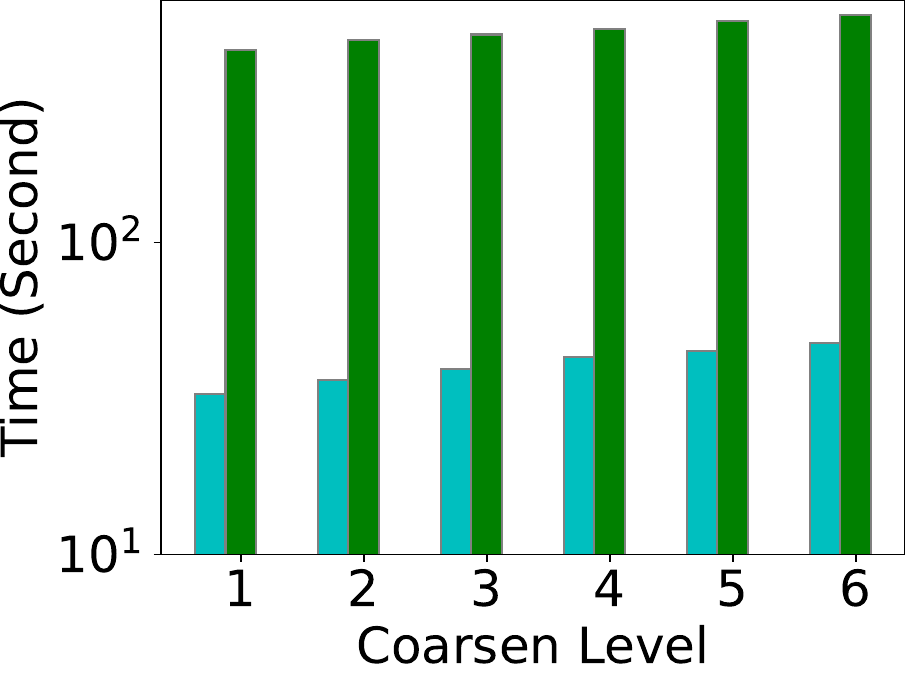}
\label{fig:academicII_comp_time}}
\subfloat[DBLP]{\includegraphics[width=0.247\linewidth]{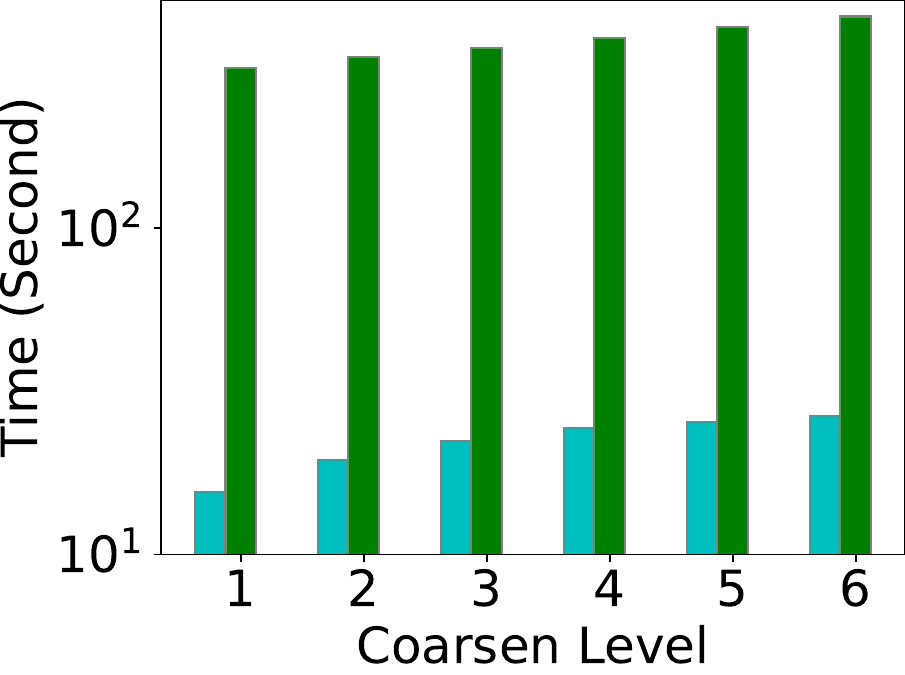}
\label{fig:dblp_comp_time}}
\subfloat[IMDB]{\includegraphics[width=0.247\linewidth]{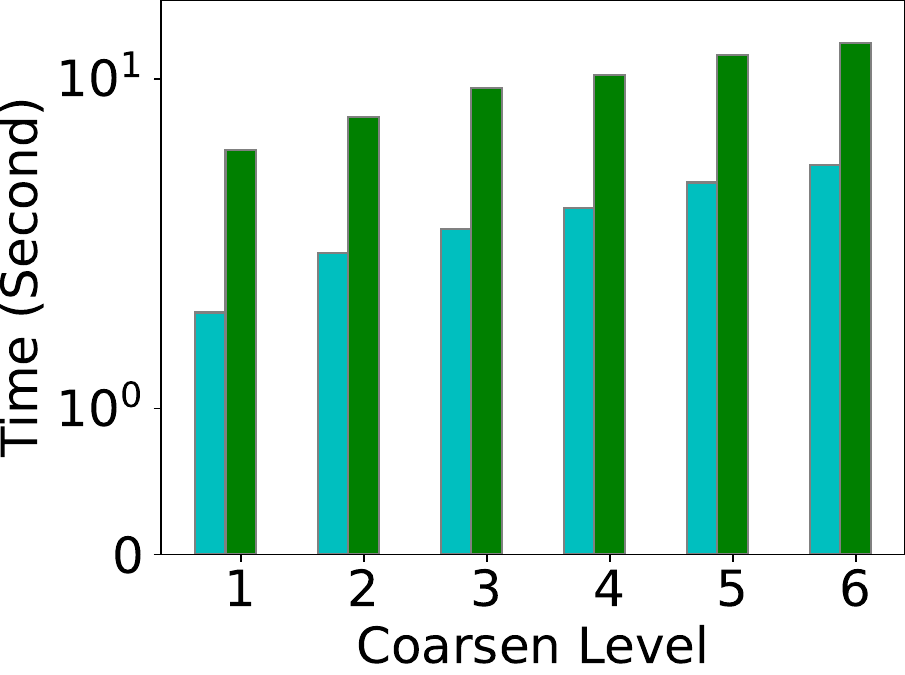}
\label{fig:imdb_comp_time}}
\subfloat[OGB\_MAG]{\includegraphics[width=0.247\linewidth]{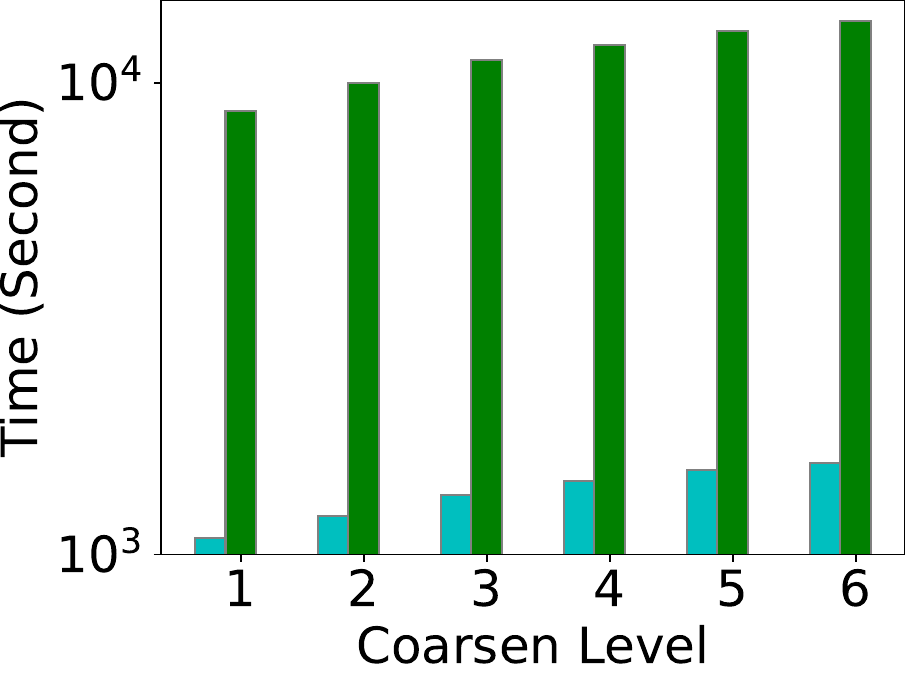}
\label{fig:ogb_comp_time}}
\caption{As the number of coarsening levels increases, the coarsening time of different coarsening strategies (LSH vs Jaccard Similarity) exhibits variations, which are measured on a logarithmic scale. "level = 0" represents the original embedding method without HeteroMILE. }
\label{coarsen}
\end{figure*}

\section{Experiment}
\subsection{Experiment Setup}{:\newline}

\textbf{Datasets: }Experiments were performed on four real-world datasets, evaluating their performance by link prediction and node classification. We used four datasets including AcademicII, DBLP, IMDB, OGB\_MAG. OGB\_MAG contains around 2M nodes and 20M edges. The dataset statistics are presented in Table \ref{dataset}

\textbf{Baselines:} We select metapath2vec \cite{dong2017metapath2vec} and GATNE \cite{cen2019representation} as baseline methods of heterogeneous graph embedding. To evaluate our HeteroMILE, we let it run in conjunction with all four coarsening policies ( Jacc\_max, Jacc\_WRS, LSH ($k$=128), and LSH ($k$=256)) on top of metapath2vec and GATNE as the base models.
% which is based on the idea of using metapaths (a sequence of edges in a graph) to define the context of a node, and then using this context to learn a low-dimensional representation of the node. 
% Below is the detailed setting and evaluation of HeteroMILE framework.

\textbf{HeteroMILE-specific Settings: }In our implementation of the HeteroMILE framework, we used an embedding dimensionality of 128 for the base-embedding method. We also experimented by varying the coarsening level, $m$, from 1 to 6 when applicable. For the heterogeneous graph convolution network model, we used four hidden layers and employed the $ELU$ activation function \cite{rasamoelina2020review}. The Adam Optimizer \cite{haji2021comparison} was employed for training the model, with a learning rate of 0.01. The training process was carried out for 200 epochs.

\textbf{Evaluation Metrics: } 
The quality of the embeddings was assessed using multi-label node classification \cite{perozzi2014deepwalk, grover2016node2vec} with F1-score and link prediction \cite{kumar2020link} with AUROC. A 10-fold cross-validation was conducted, where the embeddings were utilized as features for node classification. For link prediction, 10\% of the edges were randomly selected to construct the test dataset, which were then removed from the training data. An equal number of negative samples were added to both the training and test sets. The efficiency of the methods was evaluated by measuring the end-to-end wallclock time in seconds for all the baselines. To ensure the accuracy of the results, five runs were generated for each coarsening level, and the average Micro-F1 and average AUROC were reported.

% \textbf{Running time: } In HeteroMILE, the scalability analysis is conducted using end-to-end wallclock time, which encompasses the entirety of the process, including the training time for the refinement model. This allows for a complete assessment of the system's performance and scalability.

\begin{table}[t]
\centering
\begin{small}
\begin{tabular}{|c|c|c|c|}
      \hline
      \textbf{M}&\textbf{LSH (sec)}&\textbf{Jacc (sec)} & \textbf{speedup}\\
      \hline
      1 & 1084.419547 & 8721.358899 & 8.0424\\
      \hline
      2 & 1204.901603 & 9991.356486 & 8.2923\\
      \hline
      3 & 1338.183869 & 11214.02438 & 8.3801\\
      \hline
      4& 1429.028702 & 12040.22179 & 8.4255\\
      \hline
      5& 1509.925771 & 12869.49342 & 8.5233\\
      \hline
      6& 1562.473834 & 13524.23359 & 8.6557\\
      \hline
      % &1,939,473&21,111,007
\end{tabular}
\end{small}
\caption{\label{compare} Coarsening Time comparison between LSH and Jaccard Similarity on OGB\_MAG dataset}
\end{table}

\textbf{System Specifications: }
The experiments were performed on a Linux machine equipped with an Intel Xeon E5-2680 CPU (28 cores, 2.40GHz) and 128 GB RAM. The HeteroMILE framework was implemented in Python, adapting the original code from the authors \cite{dong2017metapath2vec, cen2019representation, yang2021interpretable} for the base embedding methods and utilizing the heterogeneous graph convolutional network model for the refinement phase. The refinement learning component was embedded using the PyTorch package.

\subsection{HeteroMILE Framework Performance}
Figure \ref{performance} and Figure \ref{performance_gatne} show the performance of HeteroMILE on various datasets and coarsening levels for link prediction and node classification using different embedding approaches of metapath2vec and GATNE. In addition, we examined various coarsening strategies and design choices for HeteroMILE for link prediction and node classification. We observed the following when considering the coarsening level of $m$=0:

\paragraph{\textbf{The scalability of HeteroMILE}} 
HeteroMILE presents a significant improvement in the efficiency of the base embedding method. This can be observed in both Figure \ref{performance} and Figure \ref{performance_gatne}, where the utilization of one level of coarsening ($m$=1) results in an approximate 2x increase in speed compared to the original base embedding method, while maintaining comparable performance in terms of quality.

By further increasing the coarsening level to $m$=6, HeteroMILE achieves an impressive speedup of approximately 20 times compared to the original base embedding method. This efficiency gain becomes particularly evident in the case of the largest dataset, OGB\_MAG. While the base embedding method would require over 7 days to complete, HeteroMILE accomplishes the execution in approximately 8 hours, exemplifying its ability to significantly expedite the graph embedding process.

\paragraph{\textbf{Impact of HeteroMILE on embedding quality: }} From Figure \ref{performance} and \ref{performance_gatne} we can see that HeteroMILE can preserve and even improve link prediction and node classification performance across all datasets. On OGB\_MAG and AcademicII with metapath2vec, it achieves an even better quality for coarsening levels $m$ = 1 and $m$ = 2. From $m$ = 0 to $m$ = 1, the Micro-F1 score increases by more than 4\%, and for $m$ = 2, Micro-F1 remains the same as the original embedding. For link prediction on the AcademicII and DBLP datasets, we can find an obvious jump over $m$ = 1 by around 5\%. In addition, we achieved 1\% to 2\% increment on AUROC after coarsening to level $m$ = 3. The better performance is even more obvious on GATNE embedding approach in Figure \ref{performance_gatne}. As we can see on most of the datasets, HeteroMILE achieves better performance on both link prediction and node classification. On the AcamdeicII dataset, HeteroMILE improves the link prediction performance by 3\% with coarsening level $m = 1$ and achieves the same performance as GATNE at coarsening level $m = 3$. HeteroMILE also accelerates among all datasets compared to the original embedding methods.

\begin{figure}[t]
\centering
\includegraphics[width=.9\linewidth]{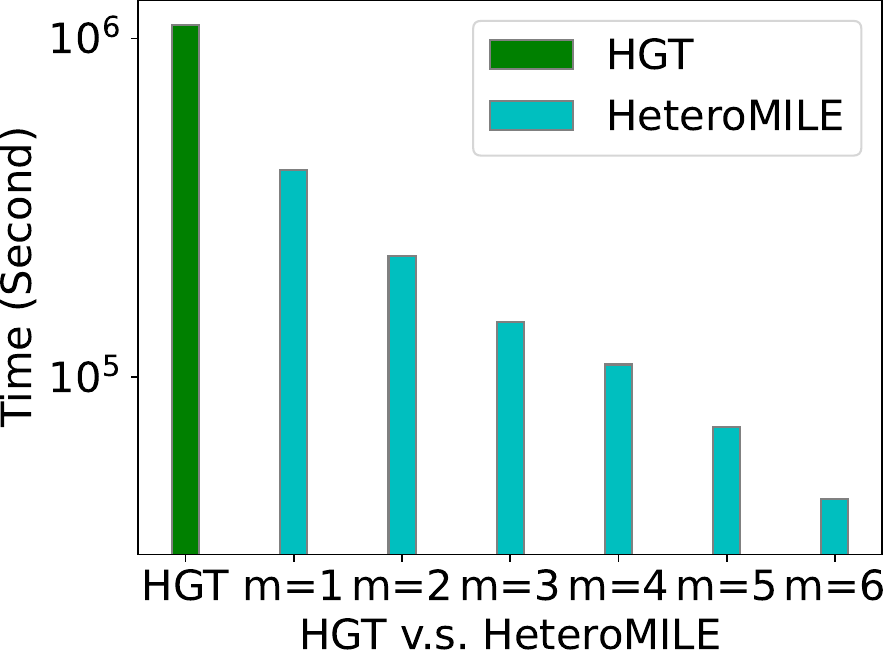}
\caption{Running time comparison of HGT versus HeteroMILE with different coarsening level $m$}
\label{baseline}
\end{figure}

\begin{figure}[]
\centering
\includegraphics[width=.9\linewidth]{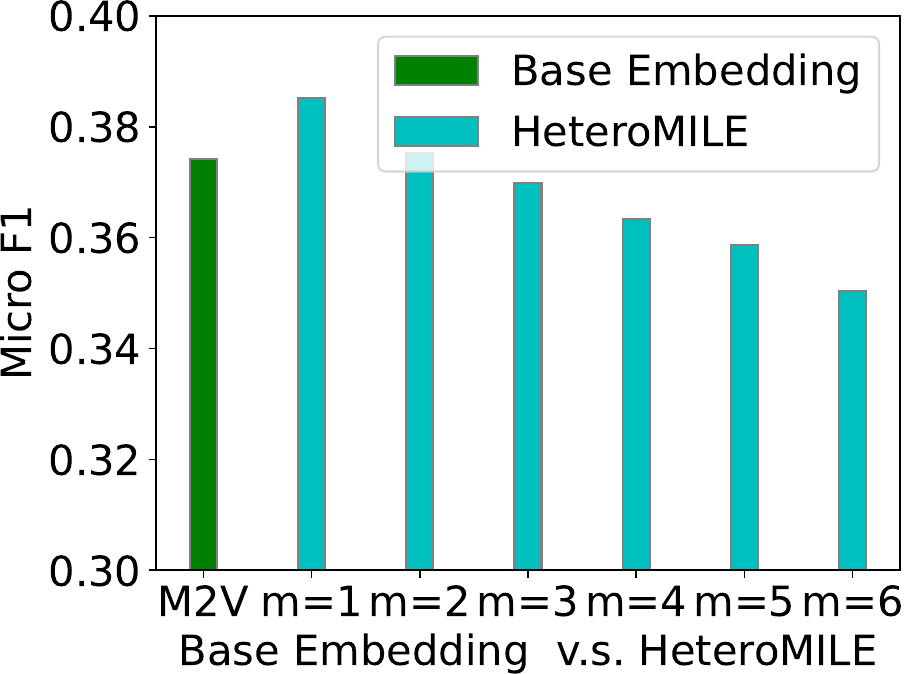}
\caption{$F_1$ score comparison of Base Embedding versus HeteroMILE with different coarsening level $m$}
\label{baseline_perf}
\end{figure}

\paragraph{\textbf{Impact of varying coarsening levels on HeteroMILE: }} As shown in Figure \ref{performance} and \ref{performance_gatne}, when coarsening level $m$ is small ($m$ = 1 and 2), HeteroMILE can preserve the performance and even improve the quality of embeddings on some datasets with a considerable reduction in time. 
In our experiments on the OGB\_MAG and AcademicII datasets, we observe about 3\% to 4\% improvement in the Micro-F1 score when using HeteroMILE with $m$ = 1 or $m$ = 2 compared to the original embeddings. Specifically, on the AcademicII dataset, employing HeteroMILE with $m$ = 1 leads to a remarkable 4.3\% increase in the Micro-F1 score. Notably, this improvement is achieved while reducing the computational time by half compared to the original embeddings. Furthermore, HeteroMILE demonstrates impressive efficiency gains, achieving more than 20x speedup while only experiencing a slight decrease in performance. These findings highlight the effectiveness of HeteroMILE in significantly enhancing both the quality and efficiency of graph embedding tasks on heterogeneous graphs.

Overall, the experiments show that HeteroMILE not only significantly reduces the time consumption of embedding generation, but also preserves and even improves the performance of link prediction and node classification.

\paragraph{\textbf{Impact of different coarsening strategies on HeteroMILE: }} HeteroMILE uses both Jaccard similarity matching and locality-sensitive hashing (LSH) matching. For the Jaccard similarity, we compared the two methods. One is to merge with the node that has maximum Jaccard similarity (Jacc\_Max), and the other is to use Jaccard-weighted random sampling (WRS), which selects the node based on its relative Jaccard similarity weight. Table \ref{coarsen_compare} shows that the Jaccard Max strategy achieves a better average Micro-F1 score compared to other coarsening approaches, whereas the result of WRS varies, but sometimes can achieve even better performance.

% Using Jaccard similarity matching strategy, we achieve better performance on some datasets with certain coarsening levels with a dramatic time reduction. 
To further minimize the computational cost, we explore another coarsening strategy, locality-sensitive hashing (LSH), which significantly reduces the coarsening time. From Table \ref{compare}, we can see that for the largest OGB\_MAG dataset, the coarsening time of LSH is approximately 8 times faster than Jaccard similarity, but with a slight decrease in performance. Therefore, we tuned the number of hashing functions, which had little influence on time but improved the performance based on Figure \ref{coarsen}. From Table \ref{coarsen_compare}, we can see that using the LSH strategy with a number of hashing functions $k$ = 256 increased the performance by 3\% than using $k$ = 128, so that the performance of LSH is close to Jaccard Similarity. 

\paragraph{\textbf{Support of multiple embedding methods on HeteroMILE: }} HeteroMILE is a generic framework allowing different embedding methods on various real-world datasets. Using metapath2vec, we can see that HeteroMILE achieves the same performance on coarsening level $m = 1$ or $m = 2$ on node classification and improves the performance on link prediction. On GATNE, HeteroMILE achieves even better results on most of the datasets. On both link prediction and node classification, HeteroMILE with coarsening level $m = 1$ improves more than 3\%. On both embedding approaches, HeteroMILE significantly speeds up while preserving the high-quality embedding performance. 

\paragraph{\textbf{Comparing HeteroMILE with HGT: }} Heterogenous Graph Transformer(HGT)\cite{hu2020heterogeneous} integrated a transformer architecture that relies on the inherent structure of the neural architecture to naturally include complex, multi-level connections between different types of nodes in the graph. This approach enables the model to autonomously discern the significance of these intricate patterns, even those that are not explicitly defined. Even though HGT improves the efficiency and efficacy of training compared to the regular GNN networks. However, it still requires a lot of memory usage to generate a promising result. Figure \ref{baseline} shows the running time comparison between HeteroMILE and HGT on the OGB\_MAG dataset. HeteroMILE using metapath2vec (M2V) as the base embedding approach with coarsening level $m = 1$  reduces the running time to half. Setting the coarsening level $m = 6$ achieves more than 20x speedup compared to HGT. Figure \ref{baseline_perf} shows that HeteroMILE improves the performance compared to the base embedding approaches of setting the coarsening level between 1 and 3. With coarsening level $m = 6$ within only 2\% performance loss, while significantly speeding up the process.

\section{Conclusion} 
In this work, we introduce HeteroMILE, a framework designed to enhance the efficiency of graph embedding methods in heterogeneous graphs. HeteroMILE achieves this by seamlessly integrating existing embedding techniques, thereby improving the scalability of the methods without requiring any modifications. By leveraging the properties of the graph and the chosen embedding method, HeteroMILE effectively reduces both the runtime and memory usage associated with the embedding process. Notably, HeteroMILE not only enhances efficiency but also often improves the quality of node embeddings. This framework's key contribution lies in its ability to reduce computational costs while preserving the graph structure, making it a valuable tool for graph embedding in heterogeneous settings.
%%%%%%%%%%%%%

% \section{Acknowledgments}

%\section{Appendices}

% If your work needs an appendix, add it before the
% ``\verb|\end{document}|'' command at the conclusion of your source
% document.

% Start the appendix with the ``\verb|appendix|'' command:
% \begin{verbatim}
%   \appendix
% \end{verbatim}
% and note that in the appendix, sections are lettered, not
% numbered. This document has two appendices, demonstrating the section
% and subsection identification method.

% \begin{acks}
% \end{acks}

\end{document}